\documentclass[letterpaper]{article} 
\usepackage{aaai2026}  
\usepackage{times}  
\usepackage{helvet}  
\usepackage{courier}  
\usepackage[hyphens]{url}  
\usepackage{graphicx} 
\usepackage{natbib}  
\usepackage{caption} 
\usepackage{algorithm}
\usepackage{algorithmic}

\usepackage{booktabs}
\usepackage{amssymb}
\usepackage{amsmath}
\usepackage{subfigure}

\usepackage{newfloat}
\usepackage{listings}
\DeclareCaptionStyle{ruled}{labelfont=normalfont,labelsep=colon,strut=off} 
\floatstyle{ruled}
\newfloat{listing}{tb}{lst}{}
\floatname{listing}{Listing}
\title{How does Chain of Thought Think?\\ Mechanistic Interpretability of Chain-of-Thought Reasoning with Sparse Autoencoding}
\author{
    Xi Chen, Aske Plaat, Niki van Stein
}
\affiliations{
    Leiden University\\

}

\usepackage{bibentry}

\begin{document}

\maketitle

\begin{abstract}

Chain‑of‑thought (CoT) prompting boosts Large Language Models accuracy on multi‑step tasks, yet whether the generated ``thoughts'' reflect the true internal reasoning process is unresolved. We present the first feature‑level causal study of CoT faithfulness. Combining sparse autoencoders with activation patching, we extract monosemantic features from Pythia‑70M and Pythia‑2.8B while they tackle GSM8K math problems under CoT and plain (noCoT) prompting. Swapping a small set of CoT‑reasoning features into a noCoT run raises answer log‑probabilities significantly in the 2.8B model, but has no reliable effect in 70M, revealing a clear scale threshold. CoT also leads to significantly higher activation sparsity and feature interpretability scores in the larger model, signalling more modular internal computation. For example, the model's confidence in generating correct answers improves from 1.2 to 4.3. We introduce patch‑curves and random‑feature patching baselines, showing that useful CoT information is not only present in the top-K patches but widely distributed. Overall, our results indicate that CoT can induce more interpretable internal structures in high-capacity LLMs, validating its role as a structured prompting method.
\end{abstract}


\section{Introduction}
While Large Language Models (LLMs) have shown exceptional performance in reasoning tasks \citep{wei2022chain}, their internal decision-making often remains a black box, making it hard for people to understand how the models reach their conclusions. 

In response to this challenge, mechanistic interpretability (MI) has emerged as a powerful alternative to traditional attributional methods \citep{chuang2024faithlm} and symbolic approaches \citep{xu2024faithful, li2024leveraging}. Instead of relying on external proxies, MI investigates how specific features, neurons, or internal circuits contribute to reasoning. However, truly "looking inside" LLMs remains challenging: classic neuron-level analyses are limited by polysemanticity \citep{bricken2023towards} and superposition \citep{elhage2022toy}, while circuit-level mapping often requires intensive manual effort, posing significant challenges for scaling to modern architectures \citep{nanda2023progress}.

A promising approach in this area is the use of sparse autoencoders (SAEs) \citep{cunningham2023sparse}. By enforcing sparsity, SAEs help resolve polysemanticity and disentangle overlapping internal representations, producing monosemantic features that can be directly probed and causally manipulated. Furthermore, compared to component-level activation patching, which can be coarse-grained and ambiguous, feature-level interventions via SAEs provide potentially more targeted and semantically meaningful control over model behavior \citep{geiger2024finding, marks2024sparse}.

Chain-of-Thought (CoT) prompting   improves LLM performance on complex, multi-step reasoning tasks \citep{wei2022chain}. However, it remains unclear whether CoT reasoning is faithful: whether the intermediate reasoning steps faithfully reflect the model's true internal decision-making process, or merely serve as plausible surface-level scaffolding. 
%
%
%
There is little feature-level, causally grounded analysis of reasoning faithfulness in LLMs, especially for math word problems requiring multi-step reasoning. 

\begin{figure*}[!ht]
    \centering
    \includegraphics[width=0.9\linewidth]{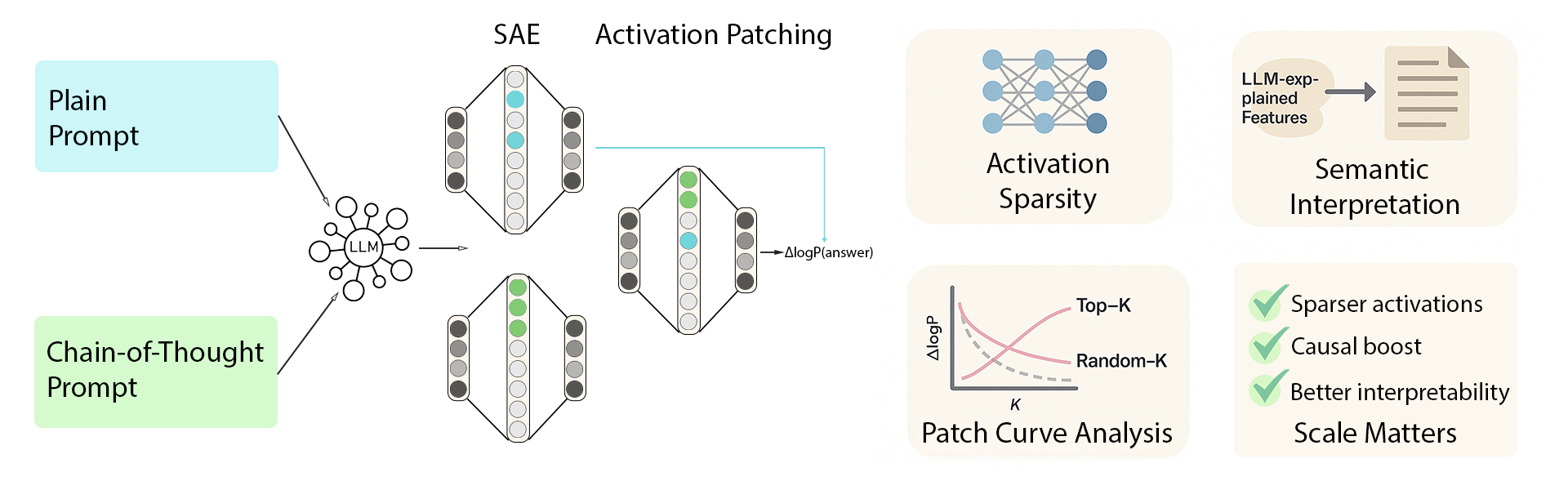}
    \caption{Workflow of the approach: After SAE, we do Activation patching, Feature Interpretation, and Activation Sparsity Analysis. All three confirm that CoT improves faithfulness of reasoning \label{fig:overview}}
\end{figure*}

To address the question whether CoT enhances faithfulness of reasoning,
we combine SAE and activation patching,  to analyze the semantic features underlying LLM reasoning. 
By (1) training separate SAEs on CoT and NoCoT activations to extract dictionary features, and (2) performing a causal intervention by patching activations to swap these features between reasoning conditions. 
To further investigate the semantic alignment of these internal features, we also perform a lightweight interpretation  that maps selected features to natural language descriptions. 
We  go beyond attributional and symbolic methods to gain deeper insight into CoT reasoning. (Code at:
\url{https://github.com/sekirodie1000/cot_faithfulness}).
%
We make the following contributions (Figure \ref{fig:overview}):

\begin{itemize}
    \item We introduce a feature-level causal intervention framework to mechanistically evaluate the faithfulness of CoT prompting in LLMs.
    \item We propose a log-probability-based evaluation procedure, enabling the systematic assessment of feature-level causal impacts in  multi-step mathematical reasoning.
    \item We demonstrate on challenging math reasoning benchmarks that CoT induces sparser and more causally effective internal features, and thus indeed enhances faithful reasoning, but only in sufficiently large models.
\end{itemize}

\section{Related Work}
We present  related work on (1)  CoT, (2) SAE and interpretation, (3) faithfulness and causal interpretability.

\paragraph{Reasoning and CoT}
CoT prompting is effective at improving  performance on complex tasks such as arithmetic and symbolic reasoning \citep{wei2022chain,plaat2024reasoning}. Zero-shot CoT shows that phrases like "Let's think step by step" can elicit coherent reasoning \citep{kojima2022large}. 
However, concerns remain: models sometimes reach correct answers despite incorrect intermediate steps \citep{yee2024dissociation}, raising doubts about the faithfulness of CoT chains. We choose GSM8K as a challenging benchmark for evaluating CoT reasoning \citep{cobbe2021gsm8k}. 
GSM8K features  complex problem structures, the answers often span multiple tokens, demanding high precision and logical coherence. 
If CoT truly reflects the model’s internal problem-solving process, then the relevant causal features should still be identifiable even in these more challenging scenarios. 

\paragraph{Sparse Autoencoder and Interpretation}
Recent work in MI uses SAEs to address  superposition and polysemanticity in network representations \citep{cunningham2023sparse}. Max activation set analysis \citep{bills2023language} and probing classifiers \citep{belinkov2022probing} 
are limited by either scale or  range of labels they use. By learning an overcomplete set of latent directions with a sparsity constraint, SAEs can break down dense model activations into monosemantic and interpretable features \citep{bricken2023towards, braun2024identifying}. Crucially, SAE-derived features are not only interpretable but also causally manipulable. 
By intervening on these feature activations we can steer model behavior.  \citet{cunningham2023sparse} used activation patching at the feature level and found that replacing or removing certain SAE features led to much larger changes in the model's outputs than 
PCA. Similarly, \citet{bricken2023towards} used logit attribution to measure feature importance and showed that individual learned features make discernible contributions to the model. With SAE, 
interventions at the feature level, targeting meaningful and sparse features, and  give more precise control of model behavior than changes made at the neuron or layer level.


\paragraph{Faithfulness and Causal Interpretability}
Faithfulness is defined as the degree to which an explanation reflects the model's true decision-making process \citep{agarwal2024faithfulness}. 
Several studies have proposed the use of counterfactual interventions to measure LLM faithfulness, such as CT/CCT frameworks \citep{atanasova2023faithfulness, siegel2024probabilities} and causal mediation analysis  \citep{paul2024making}. They  emphasize the causal relationship between model explanations and reasoning. 
\citet{matton2025walk}  introduce  interventions at the semantic concept level, elevating faithfulness analysis to a higher abstraction level.

Causal analysis tools, such as activation patching \citep{meng2022locating}, test the impact of interventions on internal activations. Unlike correlational measures they give direct evidence of causal influence. Interchange interventions \citep{geiger2021causal} and ablation find responsible components in model circuits, and recent small-scale studies were able to fully reverse-engineere transformer layers, confirming circuit functions by patching and ablation \citep{nanda2023progress}. 

Current  approaches for evaluating the causality of CoT reasoning remain limited. Some studies  attempt to intervene in the model’s internal activation space to quantify the  contribution of different parts of the reasoning chain to the final answer, making some progress in measuring faithfulness \citep{zhang2023towards, yeo2024towards}. However, because reasoning in LLMs is highly parallel and redundant, most current interventions operate at the layer or component level, which makes it challenging to pinpoint the specific features or  causal mechanisms responsible for model outputs.  Backup circuits (a  self-repair mechanism)   further complicate attribution \citep{dutta2024think}.

To address these challenges, recent work has introduced  feature-level interventions: interpretable feature directions are first extracted (e.g., via SAEs), and then directly manipulated. \citet{geiger2024finding} argue that using learned feature subspaces enables finer tracking and control of model reasoning. \citet{marks2024sparse} further construct sparse causal circuits at the feature level, showing that a small number of key features can reconstruct complex behaviors. 
\citet{makelov2023subspace} note that interventions in the feature subspace can sometimes lead to {\em interpretability illusions}, where changes in model output not necessarily originate from the intended features.  \citet{wu2024reply} argue that this phenomenon reflects the inherent property of distributed representations in neural networks, and does not prevent patching or ablation methods from revealing effective structures in complex tasks. 
We further combine SAE feature space, activation patching, and CoT prompting to 
analyze the causal mechanisms in multi-step mathematical reasoning.
By using methods such as Top-K patch curves, we provide a detailed characterization of the cumulative contribution of key features, advancing feature-level causal interpretation toward higher resolution and interpretability.

Our work goes beyond prior external and attributional approaches by directly probing LLM internal representations with mechanistic interpretability. We combine SAE-based feature extraction and activation patching to causally test whether CoT-elicited features enhance faithfulness, filling a key gap left by existing methods.

\section{Methodology}
To  evaluate whether CoT improves the internal faithfulness of LLM reasoning, 
we use a  feature-level causal analysis framework:
(1) feature extraction, (2) causal intervention, (3) structural analysis, and (4) semantic interpretation. The framework 
uses SAEs to extract semantically meaningful sparse features from model hidden states. We then apply activation patching to exchange selected features between CoT and NoCoT conditions, allowing us to examine their causal impact on model outputs. To assess whether CoT prompts induce more focused and structured computation, we compare activation sparsity across conditions. Finally, we generate natural language descriptions for SAE features and compute explanation scores to evaluate their semantic 
interpretability. This 
approach enables systematic, feature-level causal evaluation of reasoning faithfulness.
We now describe the four components of our method.

\subsection{Feature Extraction}
For feature extraction, we use sparse autoencoders to extract salient latent features from the model's hidden representations $\mathbf{x} \in \mathbb{R} ^{d_{input}}$ by learning a sparse dictionary of activation directions. Specifically, the SAE consists of an encoder $f_{enc}(\mathbf{x}) = \mathbf{h}$ that maps the high-dimensional activation $\mathbf{x}$ to a sparse feature vector $\mathbf{h} \in \mathbb{R}^k$, and a decoder $g_{dec}(\mathbf{h})$ that reconstructs the original input. To enforce sparsity, we include an L1 regularization term in the loss function. The total objective is:
$$
L_{total} = L_{recon} + \lambda \lVert h \rVert_{1}
$$
where $\mathcal{L}_{recon}$ is the reconstruction loss and $\lambda$ controls the sparsity level.

We collect a large number of residual activations under both CoT and NoCoT prompting conditions, and train two separate SAE models to obtain distinct feature dictionaries $\mathbf{D}_{CoT}$ and $\mathbf{D}_{NoCoT}$. Each input $\mathbf{x}$ is encoded into a sparse vector $\mathbf{h}$, where the nonzero dimensions indicate which semantic features are activated and their respective strengths.

By extracting features with SAEs, we transform complex high-dimensional activations into a small number of latent features with clear semantic meaning. This approach builds on prior advances in neuron interpretability; for example, \citet{cunningham2023sparse} showed that training sparse dictionaries over activations can yield semantically meaningful features that support direct intervention. Extending this line of work, our study is the first to apply SAE-based feature extraction in the context of CoT prompting.

\subsection{Causal Intervention}
For causal intervention, we analyze the causal impact of features under different reasoning conditions using the activation patching method. While activation patching has been widely adopted in neural network interpretability research \citep{heimersheim2024use}, this work is the first to systematically integrate it with the SAE feature space to evaluate the faithfulness of CoT reasoning. By combining SAEs, activation patching, and CoT prompting, we construct a feature-level causal analysis framework that enables systematic evaluation and interpretation of reasoning faithfulness. This approach addresses limitations in prior work. Specifically, prior work either focused on neurons or layers \citep{dutta2024think}, or was limited to single-step reasoning \citep{hanna2023does}. In contrast, our method provides a new tool for analyzing causal mechanisms in multi-step reasoning tasks.

Concretely, for the same math problem prompted under both CoT and NoCoT conditions, we extract hidden activations $\mathbf{x}_{CoT}$ and $\mathbf{x}_{NoCoT}$, and obtain their sparse feature representations $\mathbf{h}_{CoT}$ and $\mathbf{h}_{NoCoT}$ using the SAE encoder. Given a feature subset $S$, we construct a patched feature vector by replacing the values of $\mathbf{h}_{NoCoT}$ with those from $\mathbf{h}_{CoT}$ on the selected subset:

$$
\mathbf{h}_{patch}[S] = \mathbf{h}_{CoT}[S], \quad \mathbf{h}_{patch}[\bar{S}] = \mathbf{h}_{NoCoT}[\bar{S}].
$$

This patched feature vector $\mathbf{h}_{patch}$ is then decoded back into activation space and forwarded through the remaining layers of the model to obtain a new output.

To quantify the causal effect of the patched features, we calculate the change in log-probability assigned to the correct answer before and after patching:
$$
\Delta \log P = \log P_{patched}(answer) - \log P_{baseline}(answer).
$$

A significant increase in confidence after inserting CoT features indicates that these features play a key causal role in the reasoning process.

To assess the cumulative effect of individual features, we perform patch curve analysis: features are ranked by the absolute difference $\lvert \mathbf{h}_{CoT} - \mathbf{h}_{NoCoT} \rvert$, and the Top-K features are gradually patched in. We compute $\Delta \log P$ at each step, yielding a curve that tracks how reasoning confidence changes as more features are introduced. To control for the selection bias of Top-K features, we also introduce a Random-K baseline, where $K$ features are randomly sampled from the full feature set at each step for patching. By comparing the patch curves of Top-K and Random-K, we assess whether the causal effects are concentrated in the most differentiated features or distributed more broadly.

Although both activation patching and SAEs are existing tools, this is the first study to combine them for analyzing CoT reasoning faithfulness in mathematical tasks. Prior work often operated in raw activation or neuron space, where overlapping and polysemantic representations make interpretation difficult \citep{marks2024sparse}. By applying activation patching in the SAE-derived feature space, we enable higher-resolution and more semantically targeted interventions. Together with log-probability-based evaluation, this framework provides a precise, interpretable method for assessing reasoning faithfulness in complex multi-step reasoning tasks.

\subsection{Structural Analysis}
For structural analysis, we quantify activation sparsity to compare the internal computation focus under CoT and NoCoT conditions. Activation sparsity measures the proportion of units in a model’s hidden state that are inactive (close to zero) for a given input.

Let $x^{(l)} \in \mathbb{R}^{T \times d}$ denote the activations at layer $l$ for a sequence of length $T$ and hidden size $d$. The global sparsity for a threshold $\epsilon$ is:

$$
Sparsity(x^{(l)}) = 1 - \frac{1}{T \cdot d} \sum_{t=1}^{T} \sum_{j=1}^{d} \mathbb{I}\left[ |x^{(l)}_{t,j}| > \epsilon \right]
$$

where $I[\cdot]$ is the indicator function that returns 1 if its argument is true and 0 otherwise, and $\epsilon$ is a small positive threshold.

To enable efficient computation, especially for large models, we divide the sequence into $N$ non-overlapping chunks of size $C = T / N$. For the $i$-th chunk, the chunk-wise sparsity is defined as:

$$
ChunkSparsity_i = 1 - \frac{1}{n \cdot d} \sum_{t \in chunk_i} \sum_{j=1}^{d} \mathbb{I}\left[ |x^{(l)}_{t,j}| > \epsilon \right]
$$

Here, $chunk_i$ refers to the set of time steps belonging to the $i$-th chunk.

After calculating sparsity for all chunks, we aggregate these results to obtain the global sparsity distribution across the entire dataset. This chunk-based computation is purely technical, enabling efficient processing without changing the underlying global sparsity definition.

To our knowledge, while sparse activations are often considered a signal of improved interpretability and modularity \citep{cunningham2023sparse}, few studies have examined this in the context of CoT reasoning. Through a systematic comparison of activation sparsity under CoT and NoCoT prompting, our study is the first to reveal how prompting strategies influence the internal activation structure of the model—offering important insights into how CoT prompts reshape internal computation.

\subsection{Semantic Interpretation}
For semantic interpretation, we assign each SAE feature an interpretable explanation by collecting highly activating text snippets and using a language model to generate and simulate natural language descriptions (as in \citep{bills2023language}). The explanation's quality is evaluated by correlating predicted and true activation sequences, yielding an interpretation score. We compare the distribution of interpretation scores under CoT and NoCoT, using both statistical tests and box plots.

In our framework, the semantic interpretation module builds on prior work that uses LLMs to automatically generate feature-level semantic labels. However, we apply this technique to a novel comparative setting, analyzing differences in semantic consistency between internal features under CoT and NoCoT prompting. This perspective has not been systematically explored before. By combining explanation scores with results from explanation scores, causal patching, and activation sparsity, we gain a more comprehensive view of whether CoT prompts guide the model to learn more meaningful intermediate representations, thereby enabling a systematic evaluation of CoT faithfulness.

\subsection{Experiment Setup}
We selected two pretrained language models released by EleutherAI, Pythia-70M (6 layers, 512 hidden, 8 heads, FFN size 2048) and Pythia-2.8B (32 layers, 2560 hidden, 32 heads, FFN size 10240), both trained on the Pile and using the same vocabulary and tokenizer. We used the public Pythia v0 weights and performed only post-hoc analysis.

As our benchmarks we used GSM8K, containing grade-school level math word problems. All analyses were conducted on the training split. Two input formats were used: CoT (three fixed few-shot examples, each with detailed step-by-step solutions) and NoCoT (only the current problem). Prompts were hardcoded and identical across the dataset. Only the question was used as model input, with no ground-truth answer provided during inference; ground-truth was used only for evaluation.

To avoid bias, both formats were processed using the same pipeline, with a max input length of 256 tokens. Activations were extracted from the residual stream of layer 2 at the final token position. For both models, the training data for SAEs under CoT and NoCoT was identical except for the input format.

SAEs were trained separately for each model and prompt setting, with dictionary ratios of 4 and 8 representing lower and higher sparsity. Multiple SAE variants were trained per model/layer, with a representative subset chosen for downstream analysis.

For activation patching, we used two feature selection schemes:
\begin{enumerate}
    \item \textbf{Top-K}: the $K$sparse features with the largest absolute activation difference $|h^{(l)}_A - h^{(l)}_B|$.
    \item \textbf{Random-K}: a control variant that patches K features uniformly sampled from the full dictionary.
\end{enumerate}

For distributional analyses, we fix $K=20$. For patch-curve experiments, we vary $K \in {2, 4, 8, 16, 32, 64, 128}$, capping at 128 features. Up to 1000 problem pairs per condition were evaluated.

All model operations used HuggingFace Transformers and TransformerLens. Feature interpretation was performed using GPT-3.5-turbo on top-activating contexts. We used a single NVIDIA A100 GPU, 18 CPU cores, and 90GB of RAM. Full implementation and hyperparameter details are provided in the Appendix.

\begin{table*}[t]
\centering
\begin{tabular}{lrrrrrr}
\toprule
Model & CoT Mean & CoT Std & NoCoT Mean & NoCoT Std & $t$-stat & $p$-value \\
\midrule
Pythia-70M & 0.018 & 0.125 & 0.016 & 0.116 & 0.082 & 0.935 \\
Pythia-2.8B & 0.056 & 0.147 & -0.013 & 0.071 & 2.96 & 0.004 \\
\bottomrule
\end{tabular}
\caption{Statistical comparison of feature explanation scores under CoT and NoCoT prompts. Results are shown for Pythia-70M and Pythia-2.8B, including mean, standard deviation, and T-test statistics.}
\label{tab:interpt}
\end{table*}

\section{Results}
We analyze CoT and NoCoT reasoning in LLMs on the GSM8K dataset, evaluating feature interpretability, causal influence, and activation sparsity. All results are reported for Pythia-70M and Pythia-2.8B.

\subsection{Effect of CoT on Feature Interpretability}
\begin{figure}[h]
\centering
\includegraphics[width=0.48\linewidth]{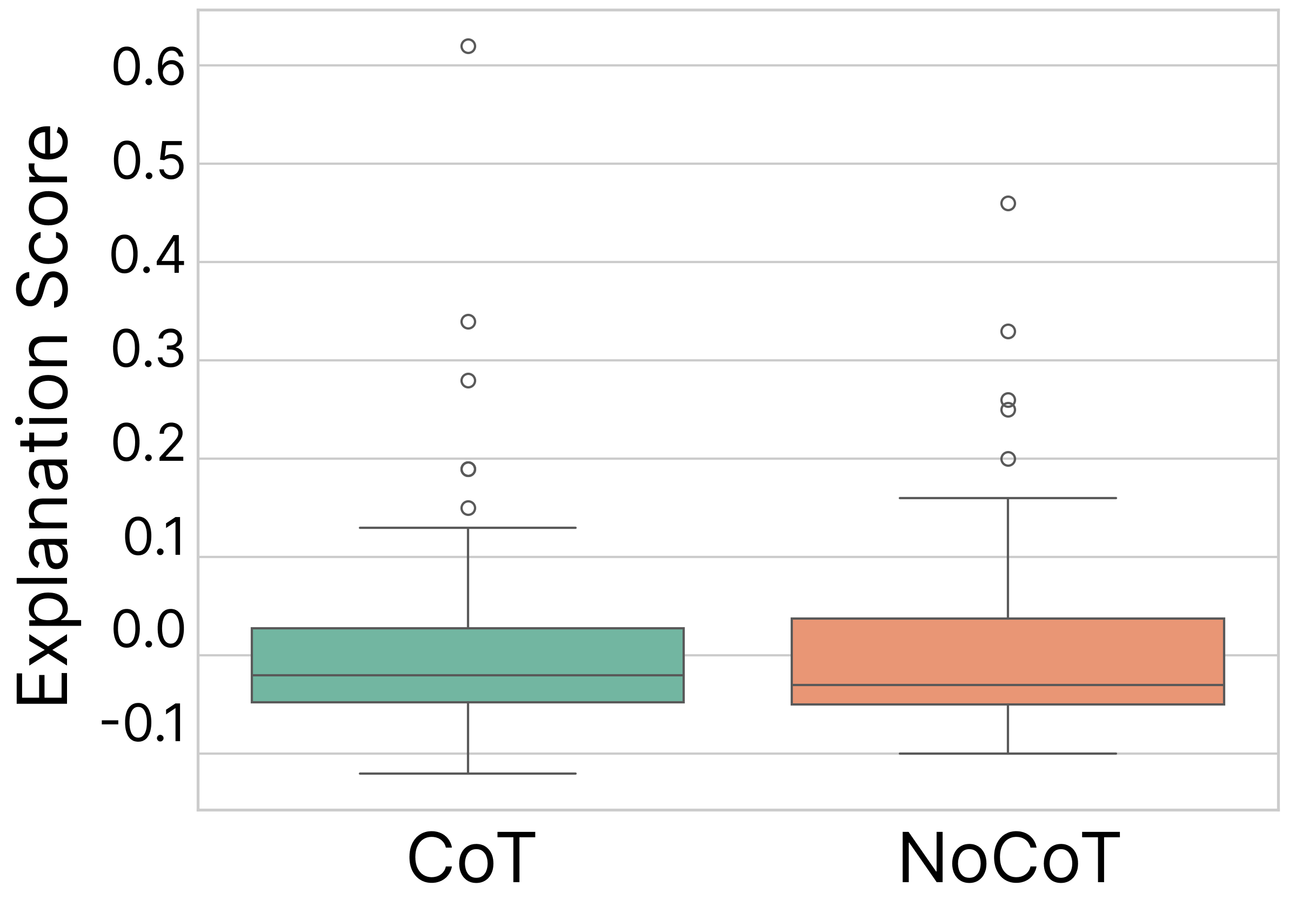}
\hfill
\includegraphics[width=0.48\linewidth]{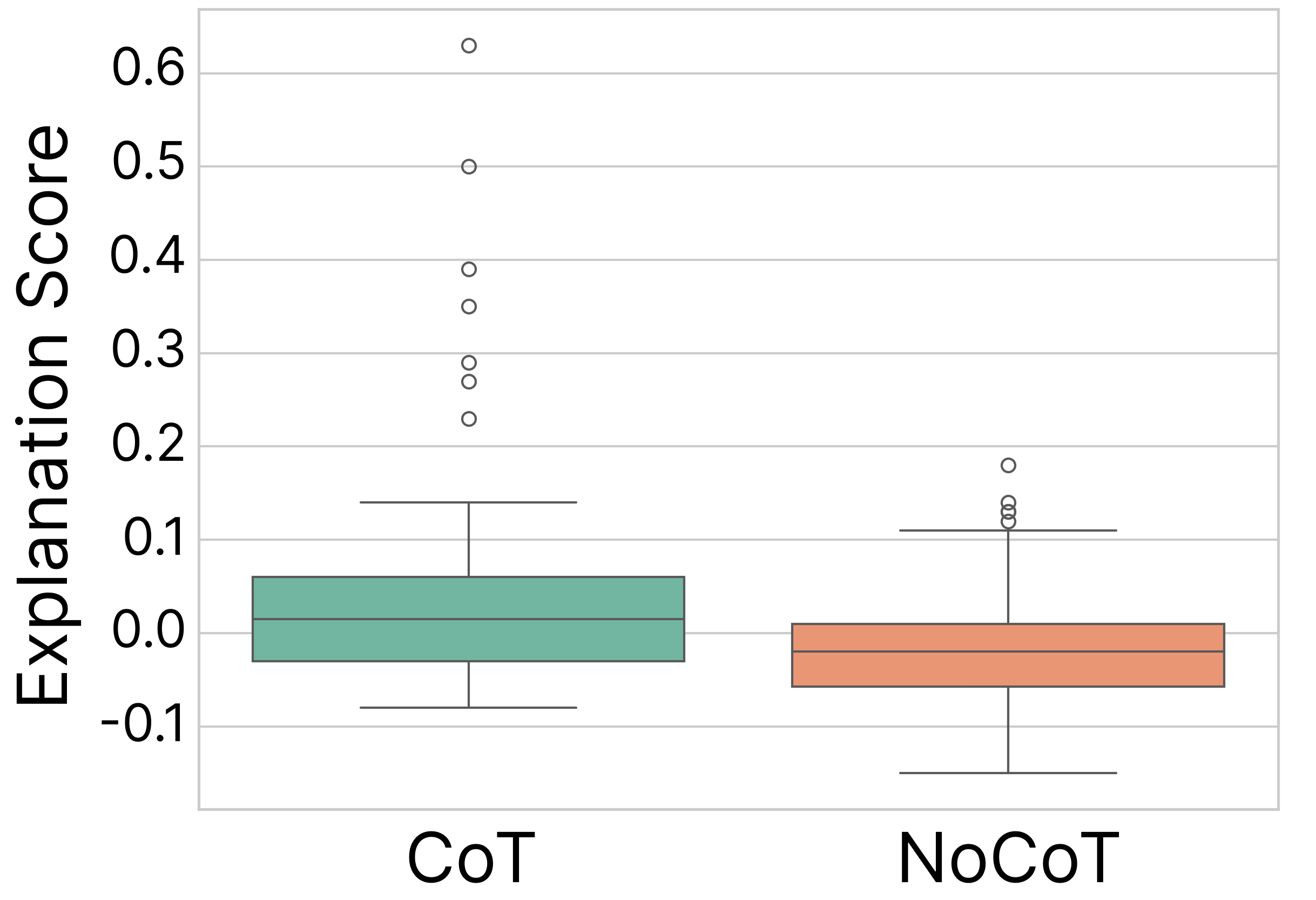}
\caption{{\small Comparison of feature explanation scores under CoT and NoCoT prompts. Left: Pythia-70M; Right: Pythia-2.8B. The 2.8B model shows higher explanation scores under CoT, indicating stronger causal features are learned in the larger model when CoT prompting is applied. Each plot is based on 50 features per condition.}}
\label{fig:feature_interpretability}
\end{figure}

We first compared the explanation scores of features learned under CoT versus NoCoT prompting. Figure \ref{fig:feature_interpretability} shows the score distributions for Pythia-70M and Pythia-2.8B under a dictionary sparsity ratio of 4. Table \ref{tab:interpt} summarizes the corresponding statistical results.
For Pythia-70M, the average explanation score under CoT was 0.018, compared to 0.016 under NoCoT,  a slight improvement. The box plot in Figure \ref{fig:feature_interpretability} further shows that features under NoCoT performed slightly better in terms of interpretability: the median score is higher and outliers are more positive. A t-test confirms this, yielding a t-value of 0.082 and a p-value of 0.935, suggesting that CoT may  slightly hinder interpretability in smaller models.
For Pythia-2.8B, the average explanation score under CoT was 0.056, higher than –0.013 for NoCoT. As shown in Figure \ref{fig:feature_interpretability}, features activated by CoT prompts display a broader distribution, with some reaching values around 0.6. This suggests that CoT  elicits semantically coherent internal features in larger models. The t-test shows this difference is statistically significant (t = 2.96, p = 0.004).

In summary, while CoT is not sufficient for logically faithful reasoning chains in LLMs, it serves as an effective structural prompt in larger models, nudging them toward more semantically coherent internal features. In smaller models, the effect remains minimal. These findings are consistent with our activation patching experiments, where CoT-elicited features in larger models demonstrated causal influence on output behavior.

\subsection{Causal Effects of CoT Features via Activation Patching}
We next examine the causal role of learned sparse features through controlled activation patching. Specifically, we inject the top-K most salient sparse features from a CoT forward pass into a NoCoT pass, and vice versa, to assess their impact on output log-probabilities for the correct answer.

\begin{figure}[t]
\centering
\includegraphics[width=0.48\linewidth]{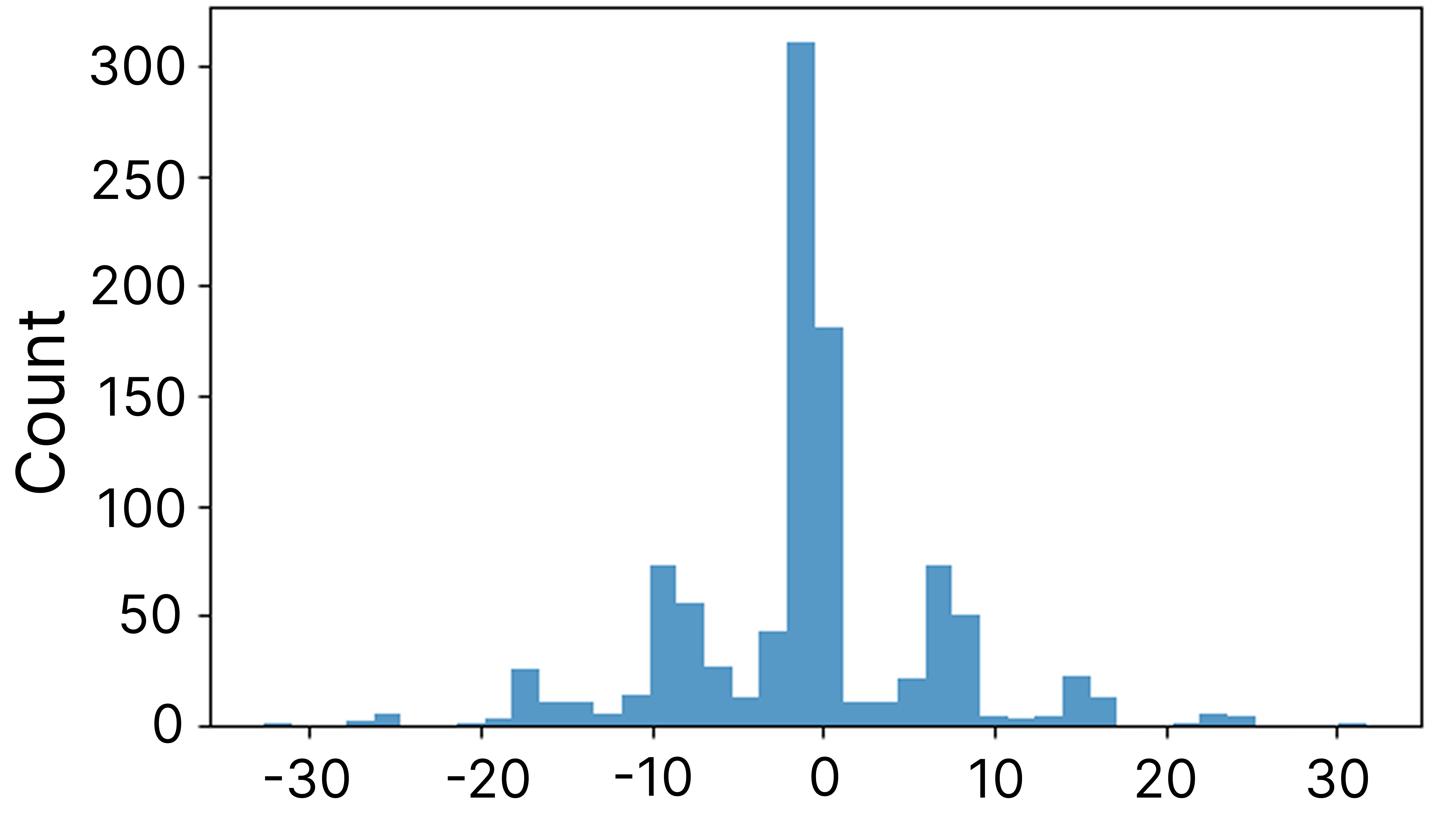}
\hfill
\includegraphics[width=0.48\linewidth]{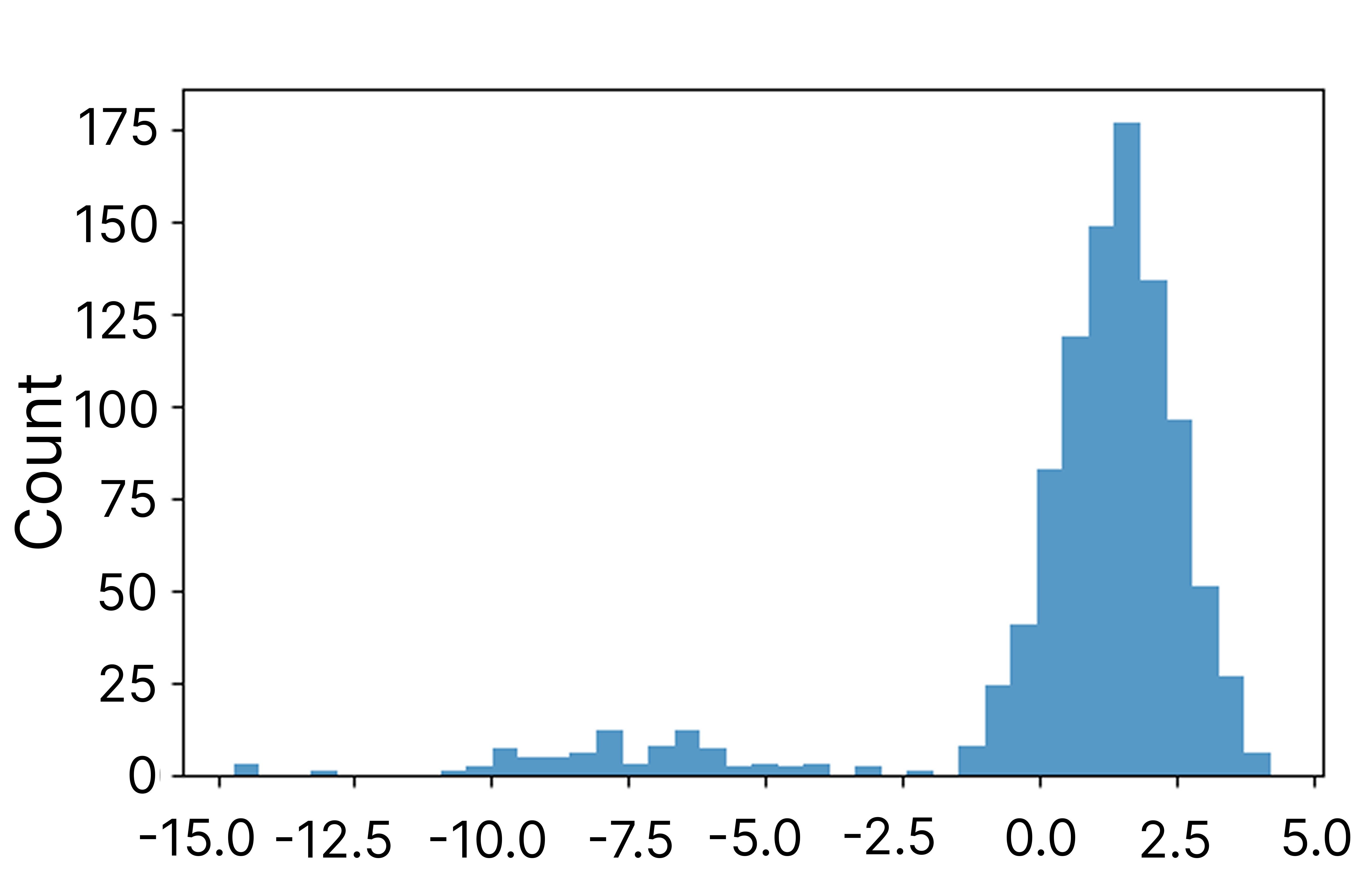}
\caption{{\small Distribution of log-probability changes after patching the top 20 CoT features into NoCoT runs under dictionary ratio 4. Left: Pythia-70M; Right: Pythia-2.8B. While 2.8B shows a strong positive shift indicating consistent benefit from CoT features, 70M shows highly variable effects, including large performance drops, suggesting unstable or less effective feature transfer.}}
\label{fig:ac_4}
\end{figure}

\begin{figure}[h]
\centering
\includegraphics[width=0.48\linewidth]{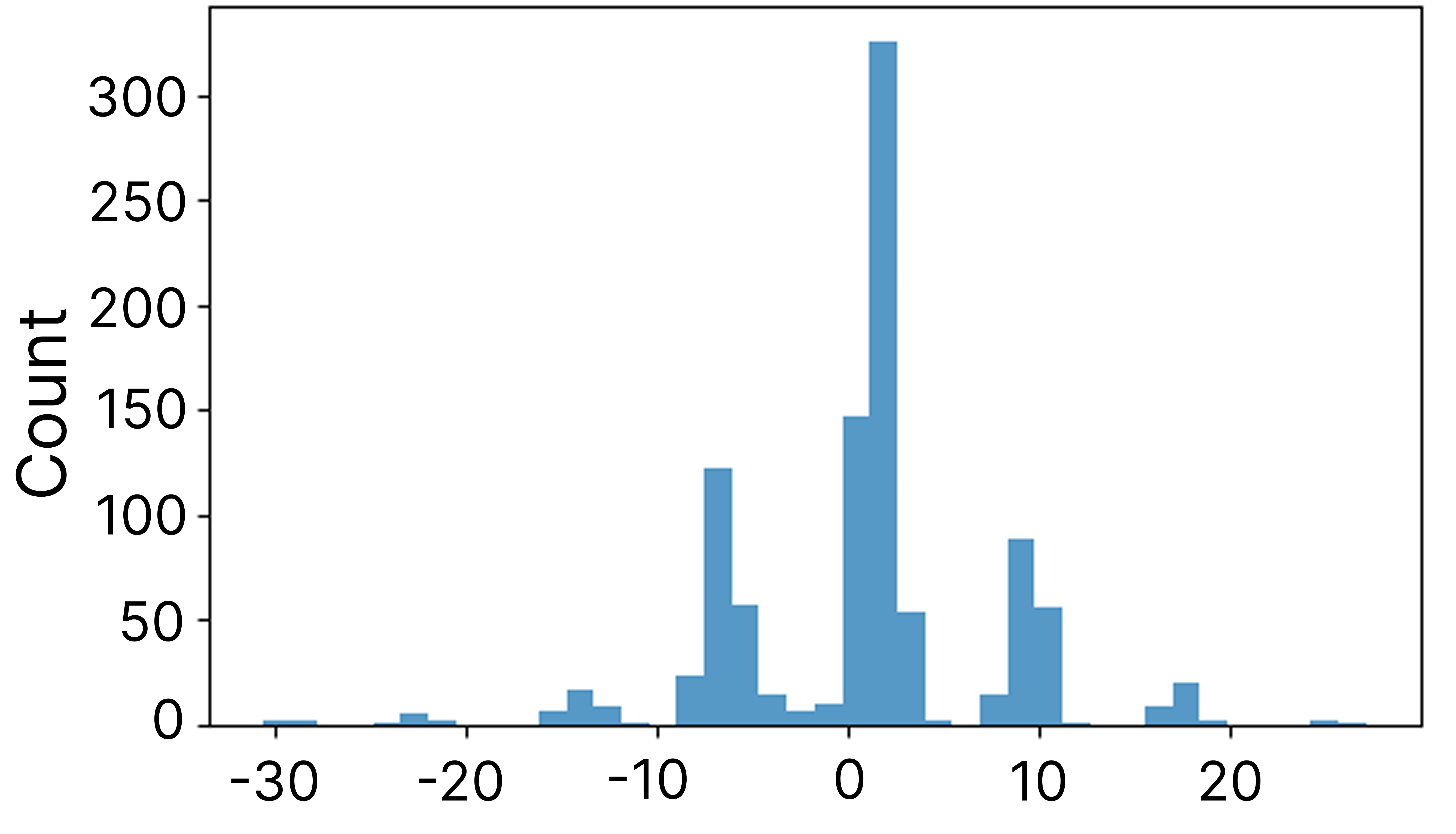}
\hfill
\includegraphics[width=0.48\linewidth]{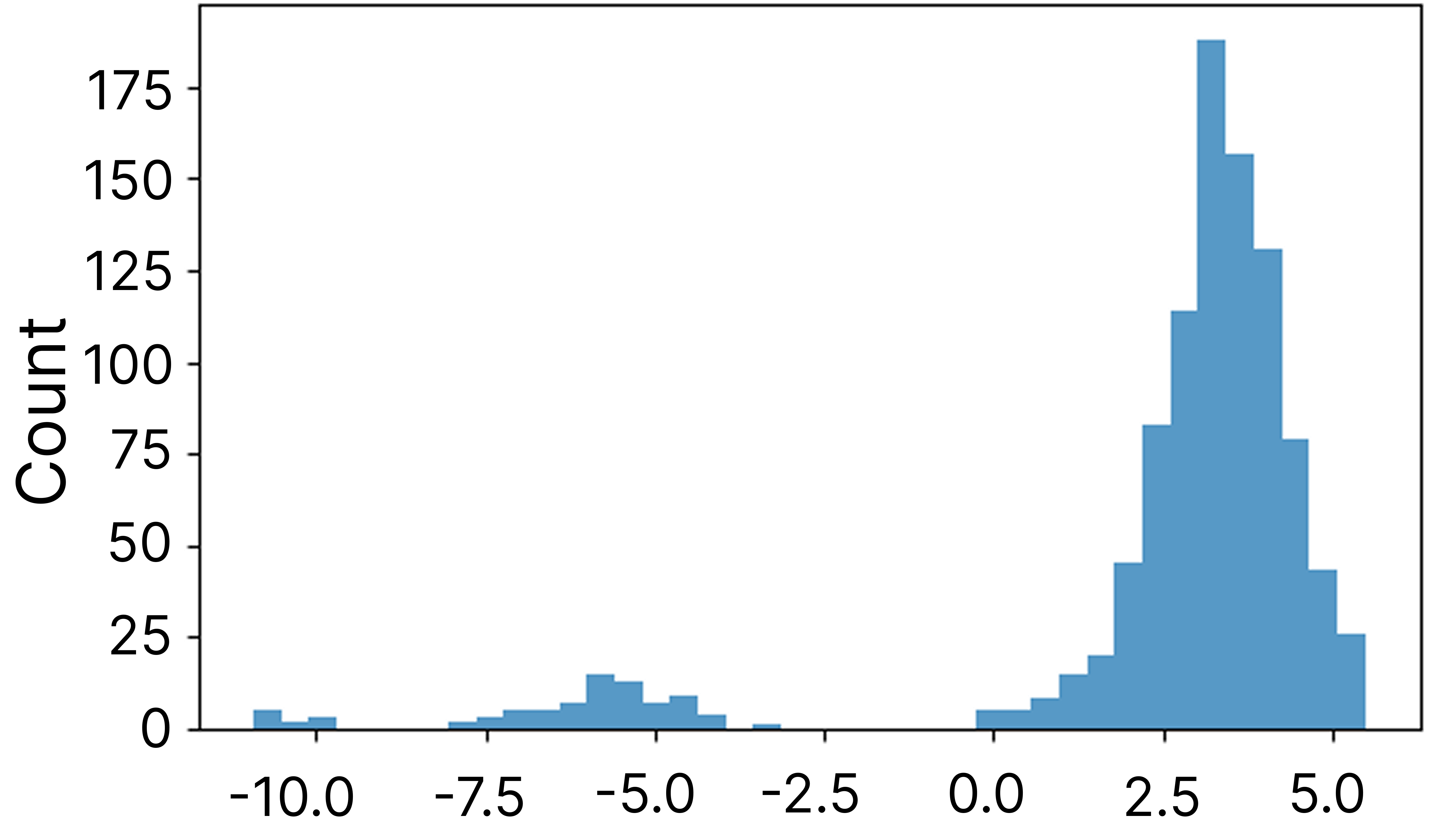}
\caption{{\small Distribution of log-probability changes after patching the top 20 CoT features into NoCoT runs under dictionary ratio 8. Left: Pythia-70M; Right: Pythia-2.8B. Compared to ratio 4, the distributions are similar: 2.8B continues to show consistent improvements, while 70M remains less robust, exhibiting high variance and frequent negative effects.}}
\label{fig:ac_8}
\end{figure}

In Pythia-2.8B, CoT-to-NoCoT patching consistently improves performance, while NoCoT-to-CoT patching has minimal effect. Figures \ref{fig:ac_4} and \ref{fig:ac_8} show that log-probability deltas after CoT patching are predominantly positive. In contrast, the same patching in Pythia-70M yields highly variable, often symmetric distributions, with both large gains and losses, indicating that CoT features do not reliably transfer in the smaller model and can disrupt original inference.

\begin{figure}[h]
\centering
\includegraphics[width=0.48\linewidth]{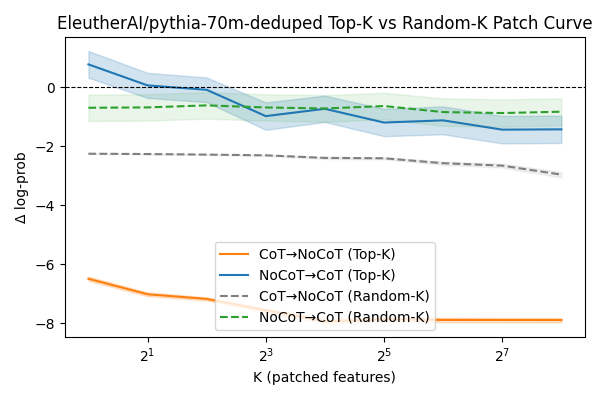}
\hfill
\includegraphics[width=0.48\linewidth]{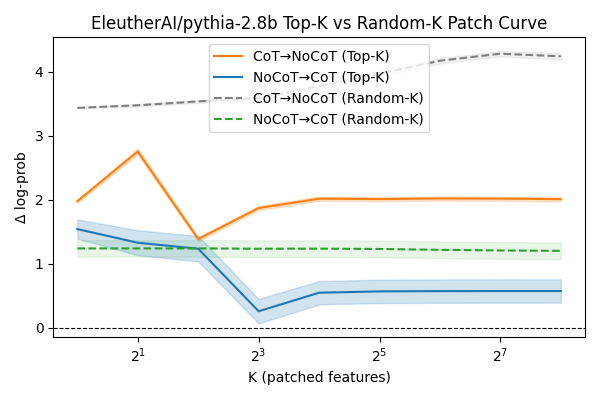}
\caption{{\small Top-$K$ and Random-$K$ patching performance under dictionary ratio 4. Left: Pythia-70M; Right: Pythia-2.8B. CoT$\rightarrow$NoCoT patching shows the effect of patching CoT features into NoCoT, while NoCoT$\rightarrow$CoT patching shows the reverse. In 2.8B, patching CoT features yields consistent performance gains, highlighting their causal importance. In contrast, for 70M, patching CoT features leads to a substantial and monotonic performance decline, suggesting that CoT-induced features are ineffective or even harmful in the smaller model ($p < 0.001$).}}
\label{fig:pc_4}
\end{figure}

\begin{figure}[h]
\centering
\includegraphics[width=0.48\linewidth]{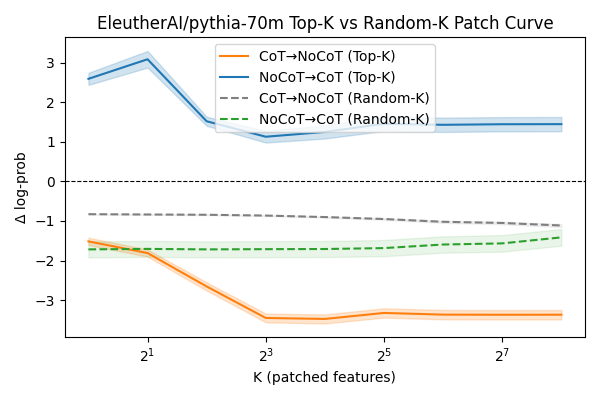}
\hfill
\includegraphics[width=0.48\linewidth]{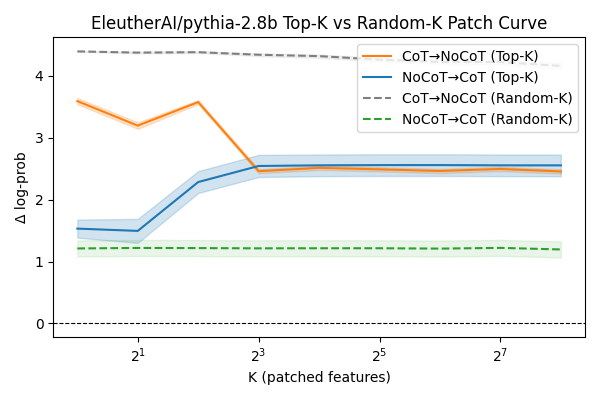}
\caption{{\small Top-$K$ and Random-$K$ patching performance under dictionary ratio 8. Left: Pythia-70M; Right: Pythia-2.8B. For 2.8B, CoT$\rightarrow$NoCoT patching consistently improves performance, with diminishing returns as $K$ increases. NoCoT$\rightarrow$CoT patching gradually degrades the CoT run, suggesting CoT features are causally significant and sparse. In contrast, for 70M, patching CoT features into NoCoT runs still causes a net performance drop, though less sharply than under ratio 4. Interestingly, NoCoT$\rightarrow$CoT patching shows mild improvement ($p < 0.001$).}}
\label{fig:pc_8}
\end{figure}

When varying the number of patched features $K$, the patching curves (Figures \ref{fig:pc_4} and \ref{fig:pc_8}) reveal that in Pythia-2.8B, injecting CoT features immediately yields a strong gain that gradually saturates, while in Pythia-70M, CoT patching leads to no gain or even performance drops. Notably, under higher sparsity (dictionary ratio = 8), these trends are even more pronounced: Pythia-2.8B's CoT curve exceeds +3.2 at K=2, then stabilizes; Pythia-70M shows persistent declines, indicating CoT features do not provide robust benefit in small models.

Crucially, random-K controlled experiments reveal that, in Pythia-2.8B, randomly sampling K CoT-activated features often outperforms selecting the Top-K by activation. For example, the model's confidence in generating correct answers improves from 1.2 to 4.3. This suggests that useful information from CoT prompts is widely distributed among moderately activated features, rather than concentrated in a few top directions. The Top-K strategy may overfit to local peaks, missing supportive features that random selection includes, resulting in more stable and comprehensive positive effects.

This distributed effect is not observed in Pythia-70M, where both random and Top-K patching fail to consistently improve performance.
This suggests that in large models, the causal signal from CoT is not limited to the most activated features, but is spread across many, making random selection more effective than simply taking the strongest activations. The next section further explains this phenomenon by analyzing the structure and sparsity of feature activations.

\subsection{Activation Sparsity under CoT and NoCoT}
\label{sec:sparsity}
Following the causal intervention experiments, we now turn to the structural properties of internal activations. We focus on sparsity—how CoT and NoCoT prompts affect the distribution and density of activated neurons and SAE features across model sizes.

\begin{figure}[h]
\centering
\includegraphics[width=0.48\linewidth]{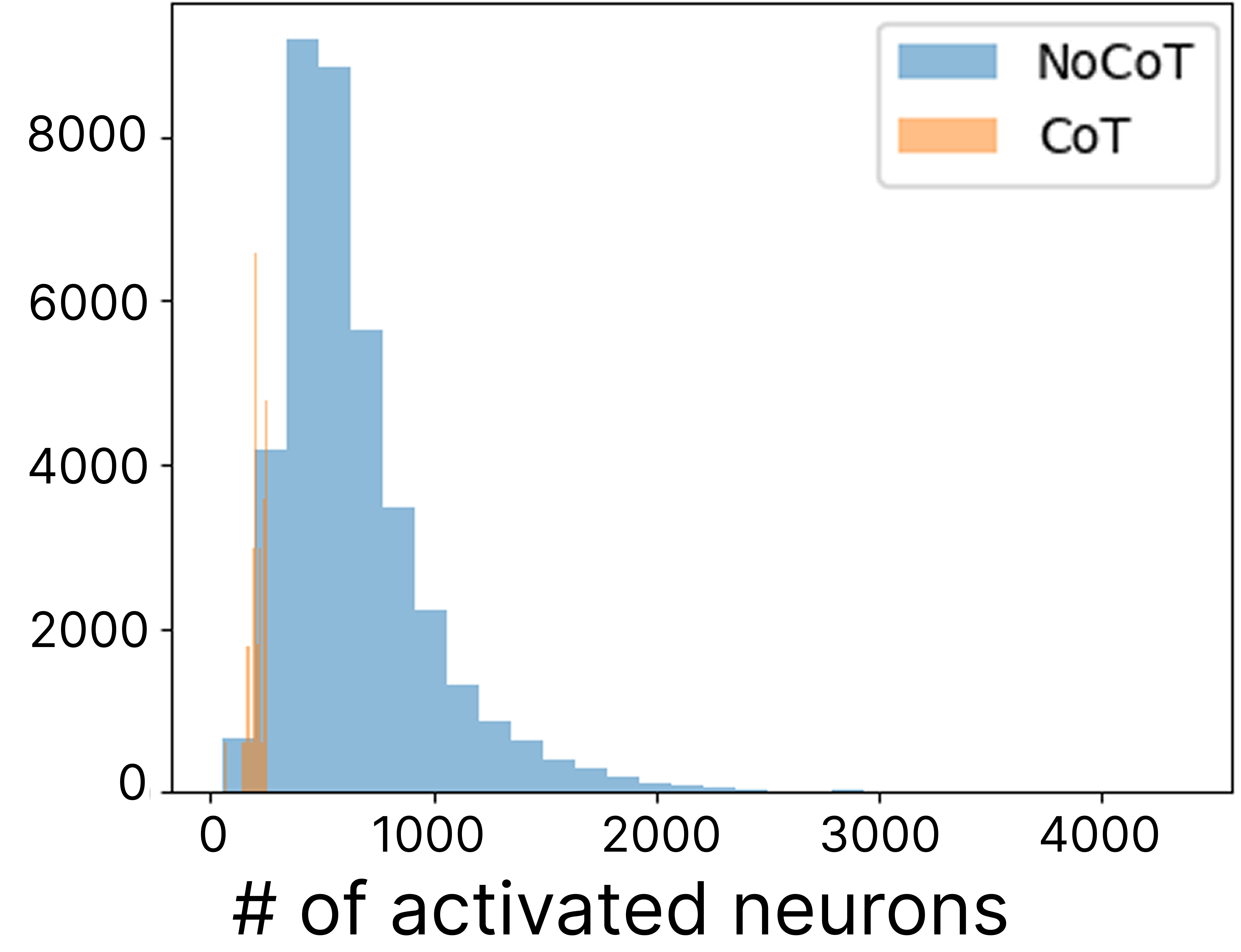}
\hfill
\includegraphics[width=0.48\linewidth]{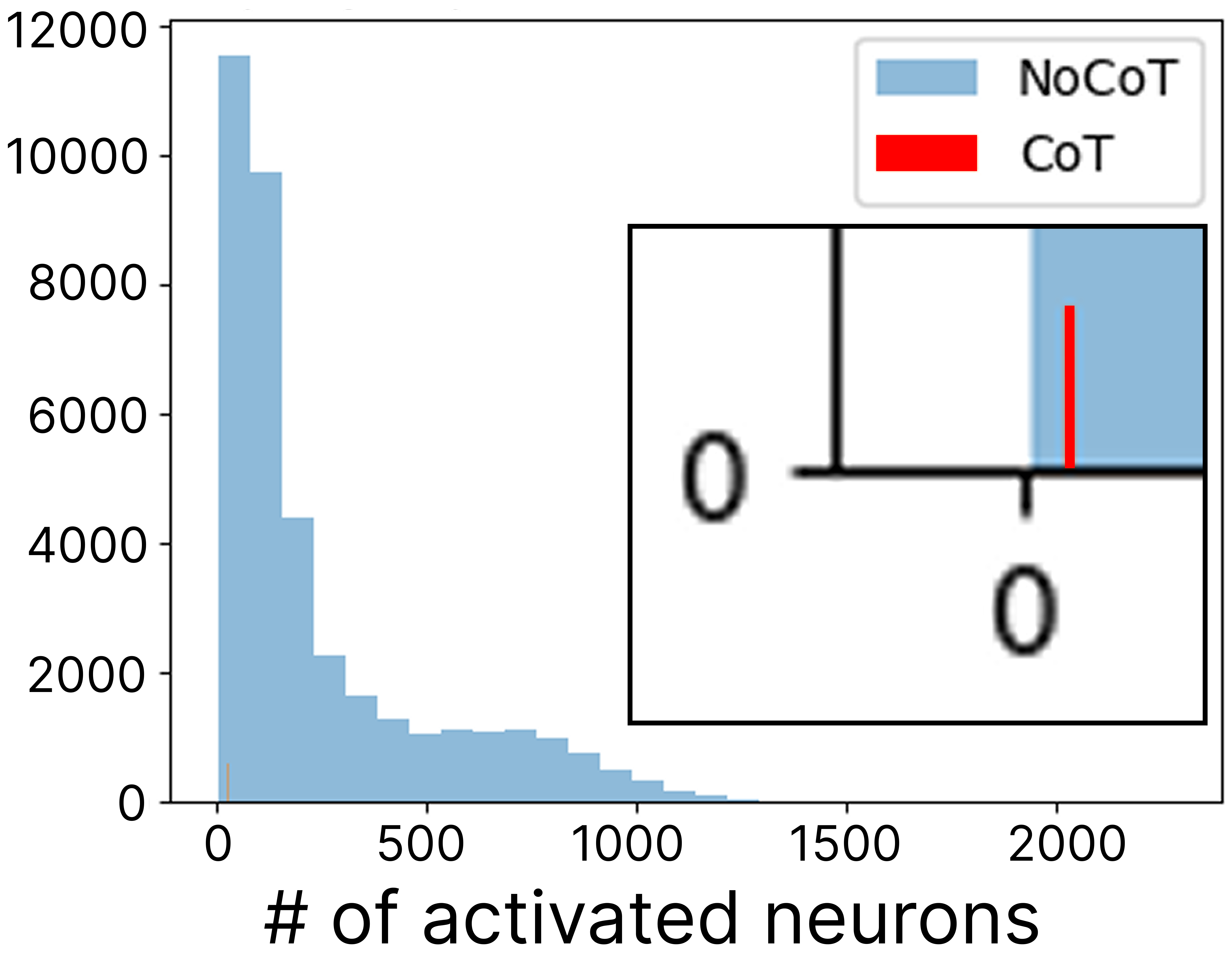}
\caption{{\small Sparsity comparison of residual activations under CoT and NoCoT prompts. Left: Pythia-70M; Right: Pythia-2.8B. In both models, CoT leads to significantly sparser residual activations, with most neurons remaining near zero and only a small subset strongly activated. This sparsity effect is markedly more pronounced in the 2.8B model, indicating enhanced activation selectivity and structured feature usage at larger scale.}}
\label{fig:sparsity}
\end{figure}

Figure \ref{fig:sparsity} shows that CoT prompts lead to significantly sparser residual activations compared to NoCoT. In the NoCoT condition, more neurons exhibit moderate to high activation; under CoT, most neurons are near zero, with only a few strongly activated. This effect is markedly more pronounced in the 2.8B model, where CoT activations are almost entirely low except for a small subset.

\begin{figure}[h]
\centering
\includegraphics[width=0.48\linewidth]{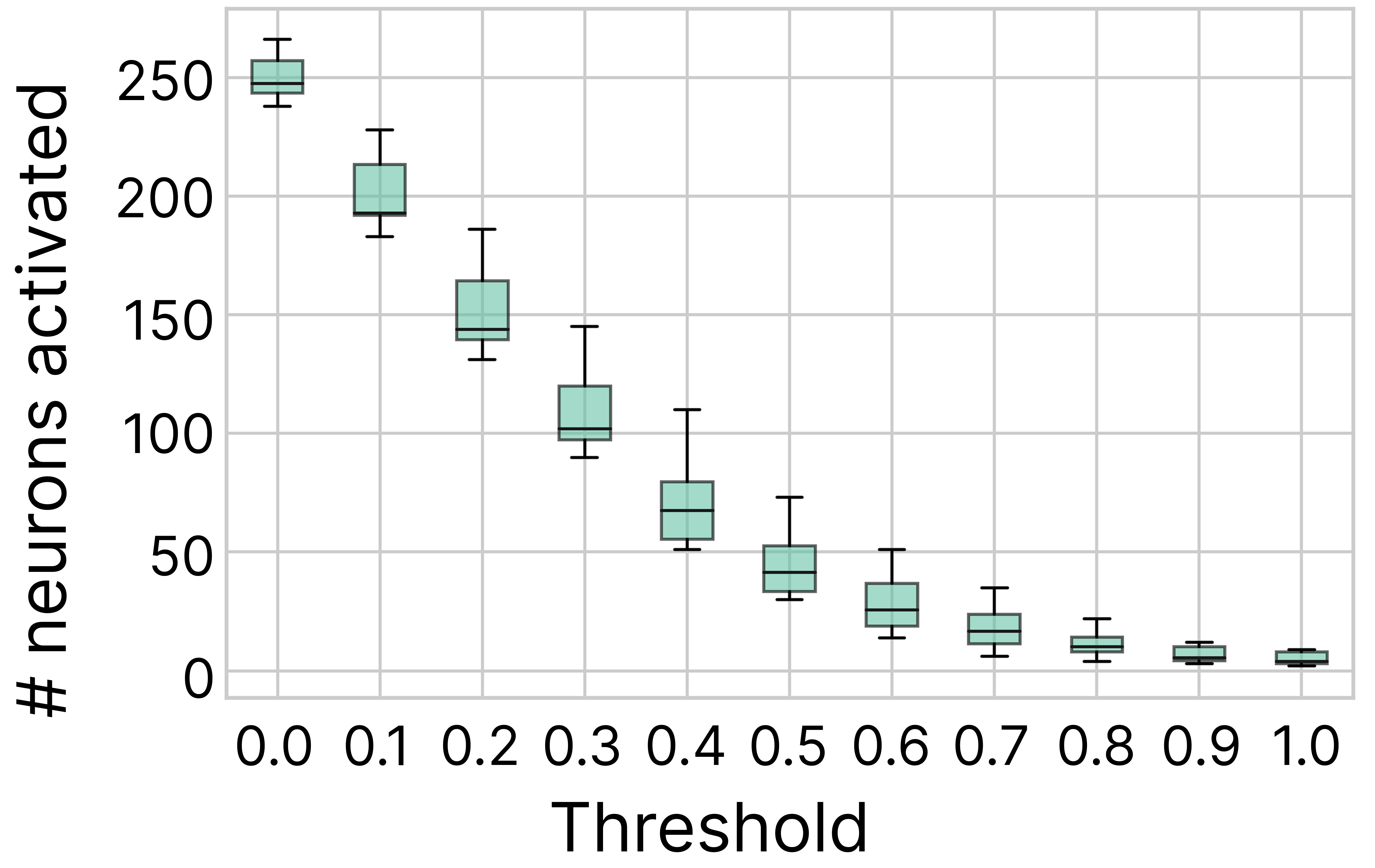}
\hfill
\includegraphics[width=0.48\linewidth]{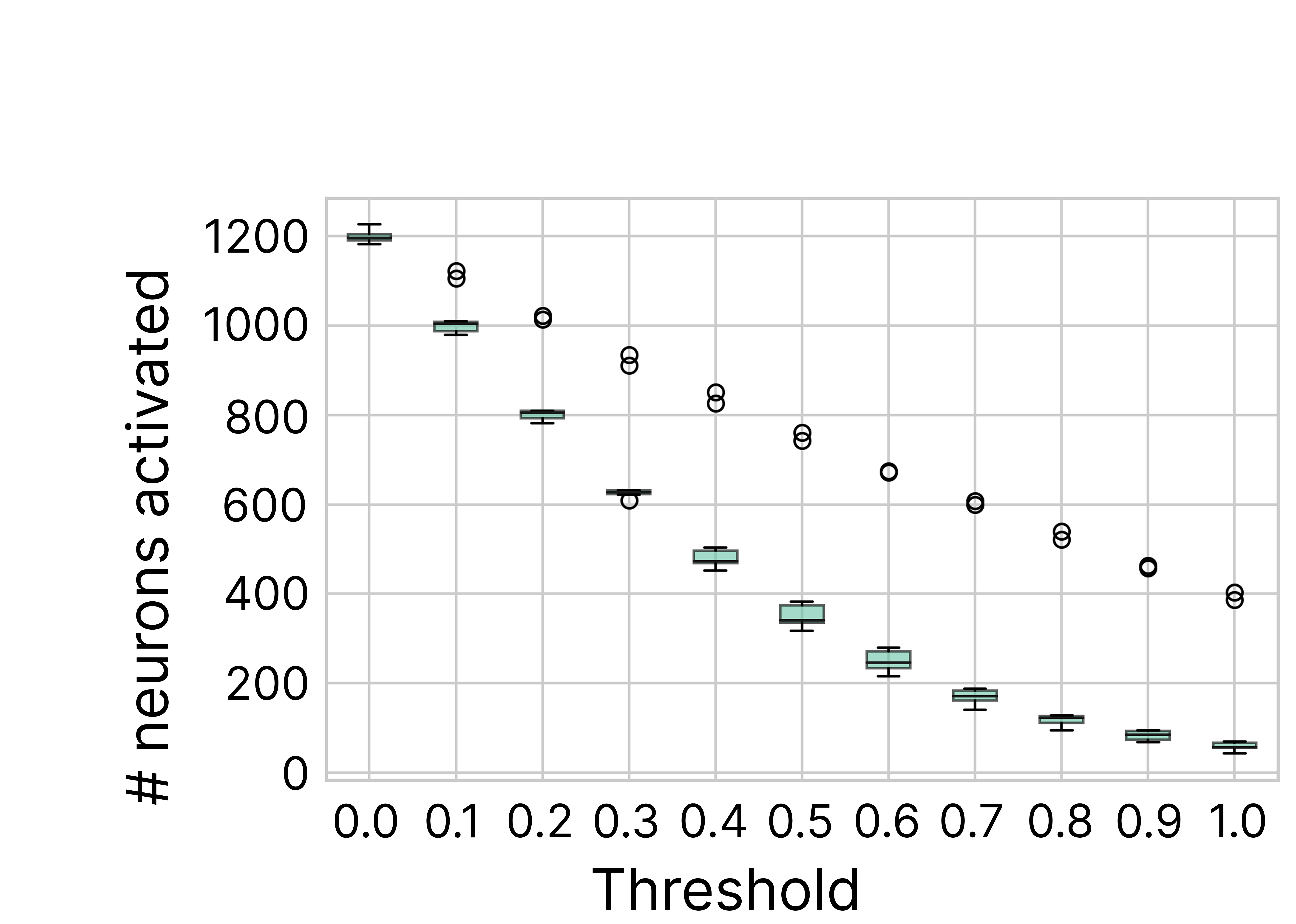}
\caption{{\small Activated neuron counts per SAE feature under NoCoT prompting, across thresholds from 0.0 to 1.0. Left: Pythia-70M; Right: Pythia-2.8B. The large model (2.8B) activates significantly more neurons per feature at each threshold, indicating denser feature composition compared to the small model.}}
\label{fig:box_nocot}
\end{figure}

\begin{figure}[h]
\centering
\includegraphics[width=0.48\linewidth]{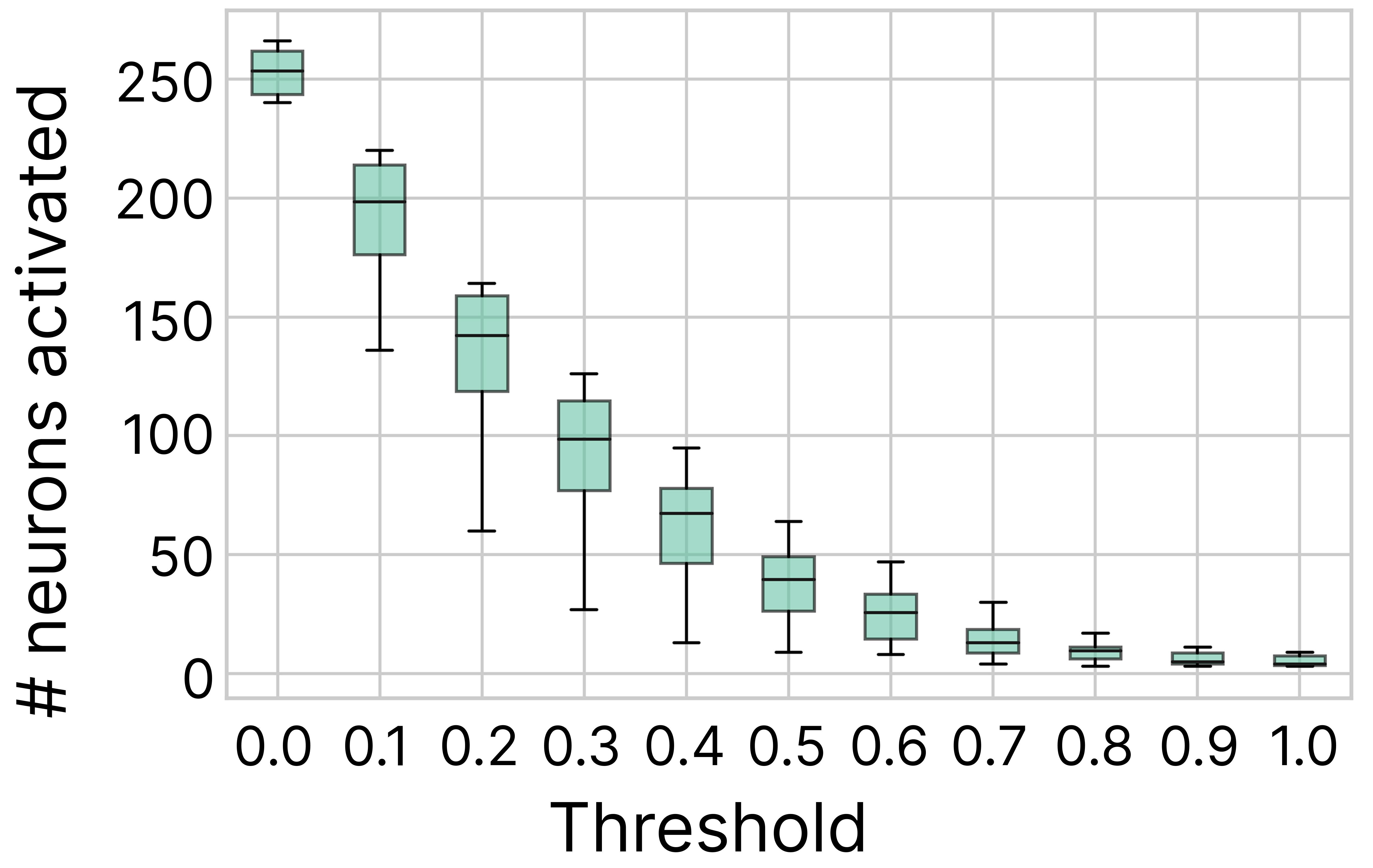}
\hfill
\includegraphics[width=0.48\linewidth]{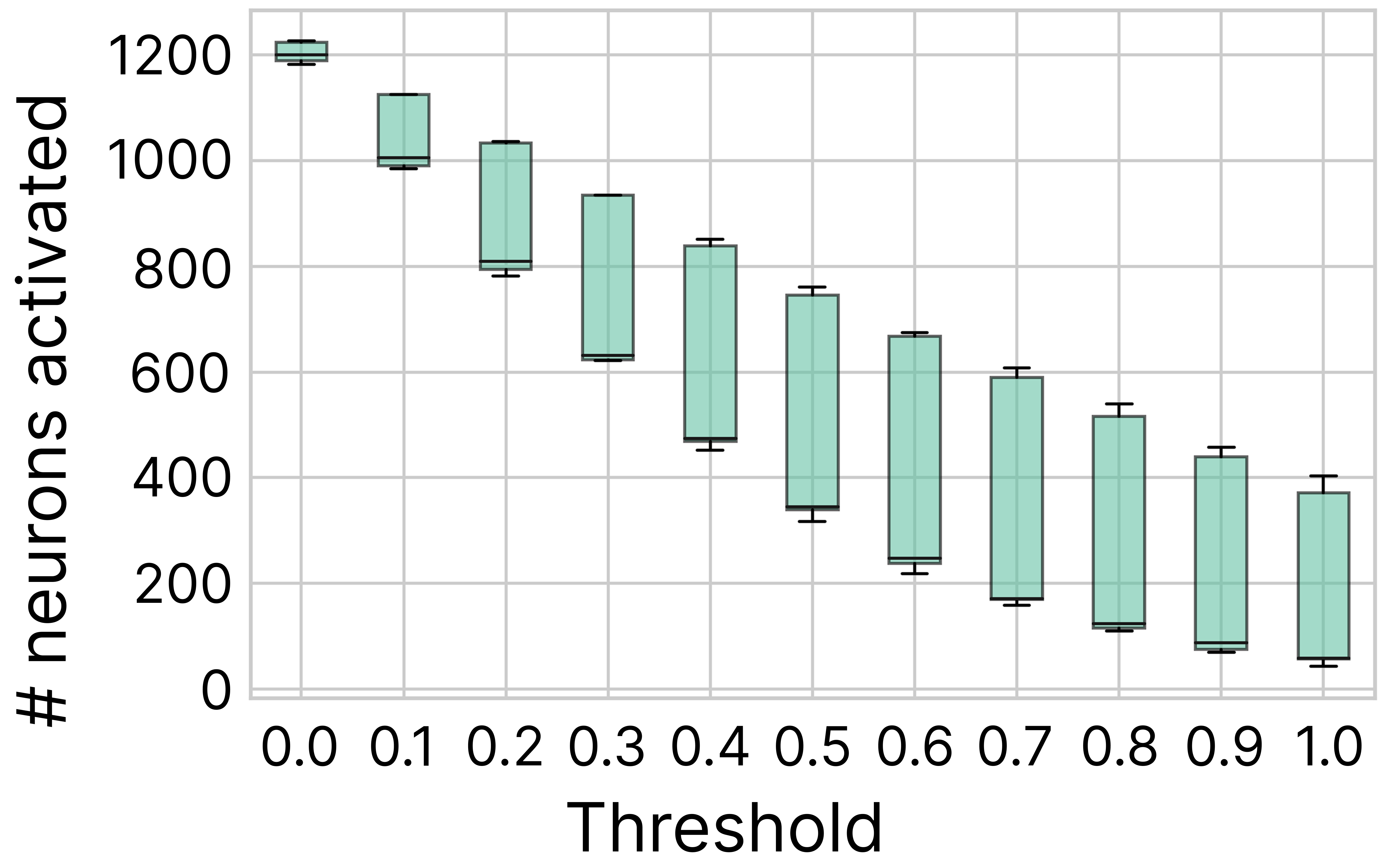}
\caption{{\small Activated neuron counts per SAE feature under CoT prompting. Left: Pythia-70M; Right: Pythia-2.8B. Compared to NoCoT, CoT prompts yield substantially sparser activations in both models, with 2.8B showing stronger sparsity and higher inter-feature variance.}}
\label{fig:box_cot}
\end{figure}

To further analyze this, we use SAE to extract feature representations and count the number of activated neurons per SAE feature. Figures \ref{fig:box_nocot} and \ref{fig:box_cot} show that under CoT, each SAE feature tends to activate only a small number of neurons, while under NoCoT, features often activate a broader set. In the 2.8B model, many CoT features are supported by only a handful of neurons, indicating a more pronounced structured sparsity.

Interestingly, this structured sparsity in CoT-induced representations also helps explain the surprising result from our patching experiments: in the 2.8B model, randomly sampled CoT features consistently outperform top-ranked ones when patched into NoCoT trajectories. At first glance, this seems counterintuitive—why would random features yield better performance than those with the highest activation? As shown earlier, CoT prompting not only increases global activation sparsity, but also leads to higher feature-level variability in the large model. Under CoT, most neurons have their activations suppressed close to zero, with only a small number strongly activated, and the number of neurons involved in different features varies greatly.

This "structured sparsity" means that the useful information activated by CoT prompts is not concentrated in a few highly activated features, but is more widely spread across many moderately activated ones. The Top-K strategy may overfit to local peaks and miss supportive features, while random sampling is more likely to include these overlooked features, leading to more stable and comprehensive positive effects in patching.

Overall, these results show that CoT prompting not only improves reasoning performance but also reshapes the model’s internal activation patterns. In both 70M and 2.8B, CoT leads to fewer neurons being activated overall, especially in the large model. At the SAE feature level, there is greater variation in how many neurons are engaged by each feature, suggesting that CoT encourages semantic resource allocation and latent disentanglement. This trend is especially prominent in 2.8B, enabling random feature patching to be surprisingly effective.

\section{Discussion}

Using mechanistic interpretability, we investigate whether CoT prompting improves the faithfulness of the reasoning processes within LLMs. Our experiments and analysis address the following three research questions:

First we studied if CoT encourages the model to learn internal features that are more semantically consistent and easier to interpret.
CoT prompts substantially indeed improve the semantic coherence and interpretability of internal features, but only in sufficiently large models. In Pythia-2.8B, features learned under CoT display higher explanation scores and semantic consistency; in 70M the effect is small.

Next we with activation patching if CoT enhances the causal relevance of internal features.
Activation patching experiments reveal a clear scale-dependent effect: in the large model, injecting random sets of CoT features into NoCoT forward passes significantly boosts output log-probabilities, demonstrating a strong causal influence. In contrast, similar interventions in the small model fail to improve, and sometimes even degrade, performance.

Finally we studied if CoT can promote sparser feature activations, a property commonly associated with enhanced interpretability.
CoT prompts induce much sparser activation patterns, especially in the 2.8B model, where both residual stream and SAE feature activations are suppressed except for a small subset. This structured sparsity enables more focused semantic allocation and explains the effectiveness of random feature patching.

\subsection{Limitation and Future Work}
This study is limited in several aspects. First, our activation patching targets only the residual activation of the final token and does not trace causal effects through the reasoning process; this is fundamentally due to the static, snapshot-based nature of the SAE framework, which is incompatible with token-level or path-level causal tracing methods \citep{goldowsky2023localizing, zhang2023towards}. Second, our interpretation module relies on OpenAI's LLM-based scoring \citep{agarwal2024faithfulness}, which offers an indirect perspective and does not ground explanations in specific neurons or heads, nor validate them with causal interventions \citep{geiger2023causal}. Third, experiments are restricted to Pythia-2.8B and smaller variants; we did not include larger models such as LLaMA-7B, and our findings may not generalize \citep{illusion-of-thinking, demircan2024sparse}. Fourth, SAE-based feature analysis introduces biases and may miss distributed or entangled representations \citep{dooms2025tokenized, karvonen2024evaluating, bereska2024mechanistic}. Not all interpretable SAE features have causal effects \citep{menon2024analyzing}.

For future work, we suggest conducting token-level and path-based causal analysis, ideally in combination with SAE-based feature decomposition, such as stepwise interventions and path patching \citep{goldowsky2023localizing}. It is important to develop activation-grounded and causally-validated explanation methods, including probing, clustering, and patching \citep{geiger2023causal, tighidet2024probing, bills2023language}. Further research should scale this framework to larger models and diverse architectures, 
and explore subspace patching and automated circuit discovery tools for more precise mechanistic analysis. 

\section{Conclusion}
This study combined sparse autoencoding, activation patching, and automated feature interpretation to probe the internal faithfulness of CoT reasoning in LLMs. Our findings show that CoT prompts, especially in larger models, induce more semantically coherent, causally effective, and sparser internal features. However, these effects are minimal in small models. While limitations remain, this work highlights how CoT not only improves output but also reshapes internal reasoning processes, offering new insight into the mechanisms underlying LLM reasoning.


\bigskip
\bibliography{aaai2026}

\appendix
\section{Appendix to: How does Chain of Thought Think?\\ Mechanistic Interpretability of Chain-of-Thought Reasoning with Sparse Autoencoding}
The main paper provides an essential summary of our experimental results, shortened due to space restrictions. We believe the main contribution lies in our analysis and therefore, in this appendix we provide 6 pages with more details of our results, as well as links to code and configuration files, for reproducibility. For readability and continuity, some passages from the main paper are repeated. We also maintain figures from the main paper but much larger.

\section*{Experimental Setup and Implementation Details}
In this study, we selected two pretrained language models released by EleutherAI, Pythia-70M and Pythia-2.8B, as our primary subjects of analysis. Pythia-70M is a small model with 6 Transformer layers, 512 hidden dimensions, and 8 attention heads, with a feedforward hidden size of approximately 2048. In contrast, Pythia-2.8B is a large-scale model with 32 layers, a model width of 2560, 32 attention heads, and a feedforward size of roughly 10240. Both models share the same vocabulary and tokenizer, and are trained on the Pile dataset. We used the publicly available weights from the Pythia v0 release, and all experiments were conducted on these frozen models, purely for post-hoc analysis and intervention.

We employed the GSM8K dataset as the benchmark for evaluating the reasoning capabilities of the language models. GSM8K contains grade-school level math word problems, each comprising a natural language question (typically one to two sentences) and a final numerical answer \citep{cobbe2021gsm8k}. All our analyses and activation collection experiments were conducted on the training split, as it includes ground-truth answers, while the test split remains hidden.

To investigate the effects of CoT prompting on model behavior, we created two distinct input formats for each problem: one following the CoT format, and the other following a NoCoT format. The CoT input consists of a fixed few-shot prompt followed by the current problem. The prompt contains three representative examples, each with a detailed step-by-step solution, followed by the current question prefixed with $\verb|Q: ... \nA:|$. These examples are fixed across the dataset, functioning as a hardcoded prompt template rather than a dynamic in-context learning setup. In contrast, the NoCoT input includes only the current problem with no examples or step-by-step guidance.

Importantly, we only used the question portion of each example as model input, without providing the ground truth answer. During inference, the model must generate a solution based solely on the given problem (and prompt, if applicable). Ground truth labels were used only for downstream evaluation. Additionally, we avoided input truncation by tokenizing the full problem statement, subject only to a maximum input length (e.g., 256 tokens). For activation recording, we ensured that both CoT and NoCoT samples were handled using the same formatting pipeline to avoid introducing bias.

For training the SAE and conducting activation-based comparisons, we applied the two input formats to the full GSM8K training set. That is, the amount of training data used in both CoT and NoCoT settings was identical, with the only difference being the input formatting. This controlled setup enables a fair comparison across reasoning modes, particularly in terms of residual activation sparsity, causal response to interventions, and structure of learned features.

All model loading, tokenization, inference, and intermediate activation extraction were implemented using the HuggingFace Transformers library and the TransformerLens interpretability toolkit. All experiments were performed on compute nodes equipped with a single NVIDIA A100 GPU, 18 CPU cores, and 90GB of RAM. We extracted activations from the residual stream of layer 2 in both models. For each forward pass, we recorded the activation at the final token position, which served as the input for feature extraction and patching experiments.

To control for dictionary sparsity and feature capacity, we trained SAE models with different dictionary ratios, specifically 4 and 8, representing lower and higher sparsity settings, respectively. For each model and layer, multiple SAE variants were trained, and a representative subset was selected for downstream interpretation and intervention experiments. 

During patching and evaluation, we considered two feature-selection schemes:

\begin{enumerate}
    \item \textbf{Top-K}: the $K$sparse features with the largest absolute activation difference $|h^{(l)}_A - h^{(l)}_B|$.
    \item \textbf{Random-K}: a control variant that patches K features uniformly sampled from the full dictionary.
\end{enumerate}

For distributional analyses, we fix $K=20$. For patch-curve experiments, we vary $K \in {2, 4, 8, 16, 32, 64, 128}$, capping the number of patched features per sample at 128 to balance signal strength and computational cost. We evaluate up to 1000 problem pairs per condition to ensure statistical power while maintaining feasibility.

This experimental design allows us to systematically analyze the behavior of internal features under explicit reasoning conditions, and to uncover how semantic representations are structured and recombined within the sparse activation space of pretrained language models.

\section*{Extended Analysis of Results}
\subsection*{Causal Effects of CoT Features via Activation Patching}
\begin{figure*}[h]
\centering 
\subfigure[Pythia70m]{
\label{70m_ac_4}
\includegraphics[width=0.4\textwidth]{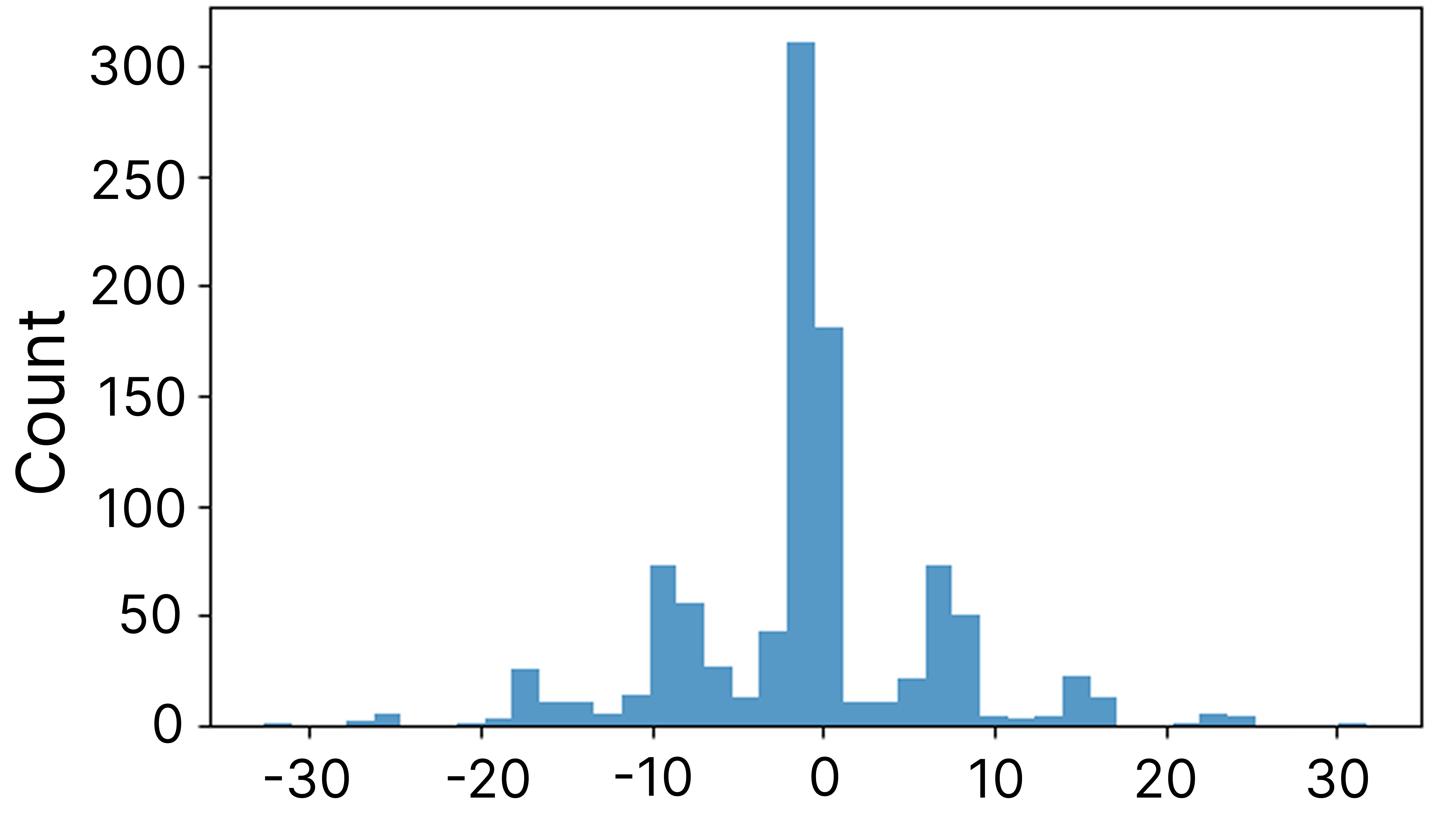}}~\subfigure[Pythia2.8b]{
\label{2.8b_ac_4}
\includegraphics[width=0.4\textwidth]{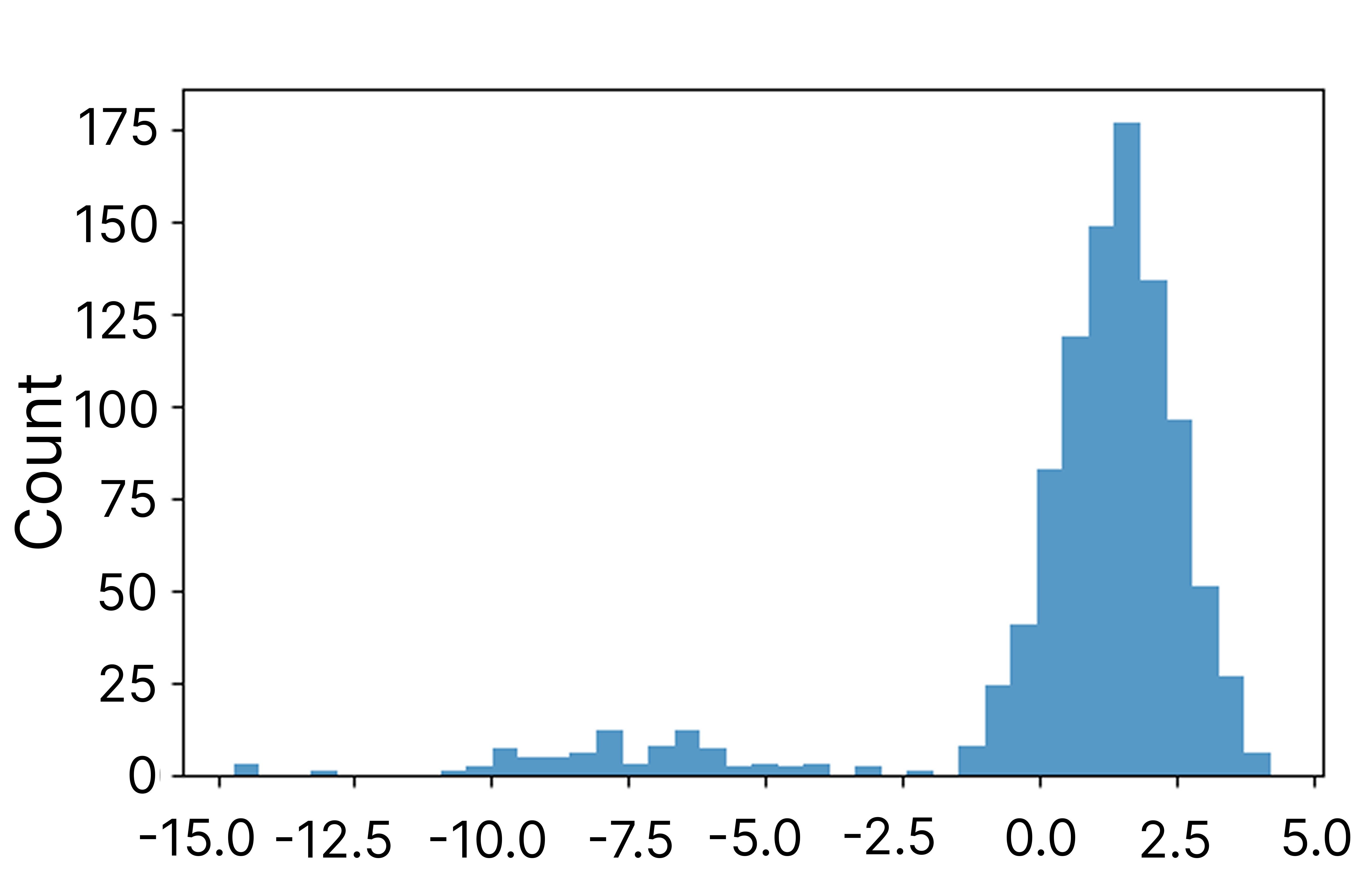}}
\caption{Distribution of log-probability changes after patching the top 20 CoT features into NoCoT runs under dictionary ratio 4. Left: Pythia-70M; Right: Pythia-2.8B. While 2.8B shows a strong positive shift indicating consistent benefit from CoT features, 70M shows highly variable effects, including large performance drops, suggesting unstable or less effective feature transfer.}
\label{2}
\end{figure*}

\begin{figure*}[h]
\centering 
\subfigure[Pythia70m]{
\label{70m_ac_8}
\includegraphics[width=0.4\textwidth]{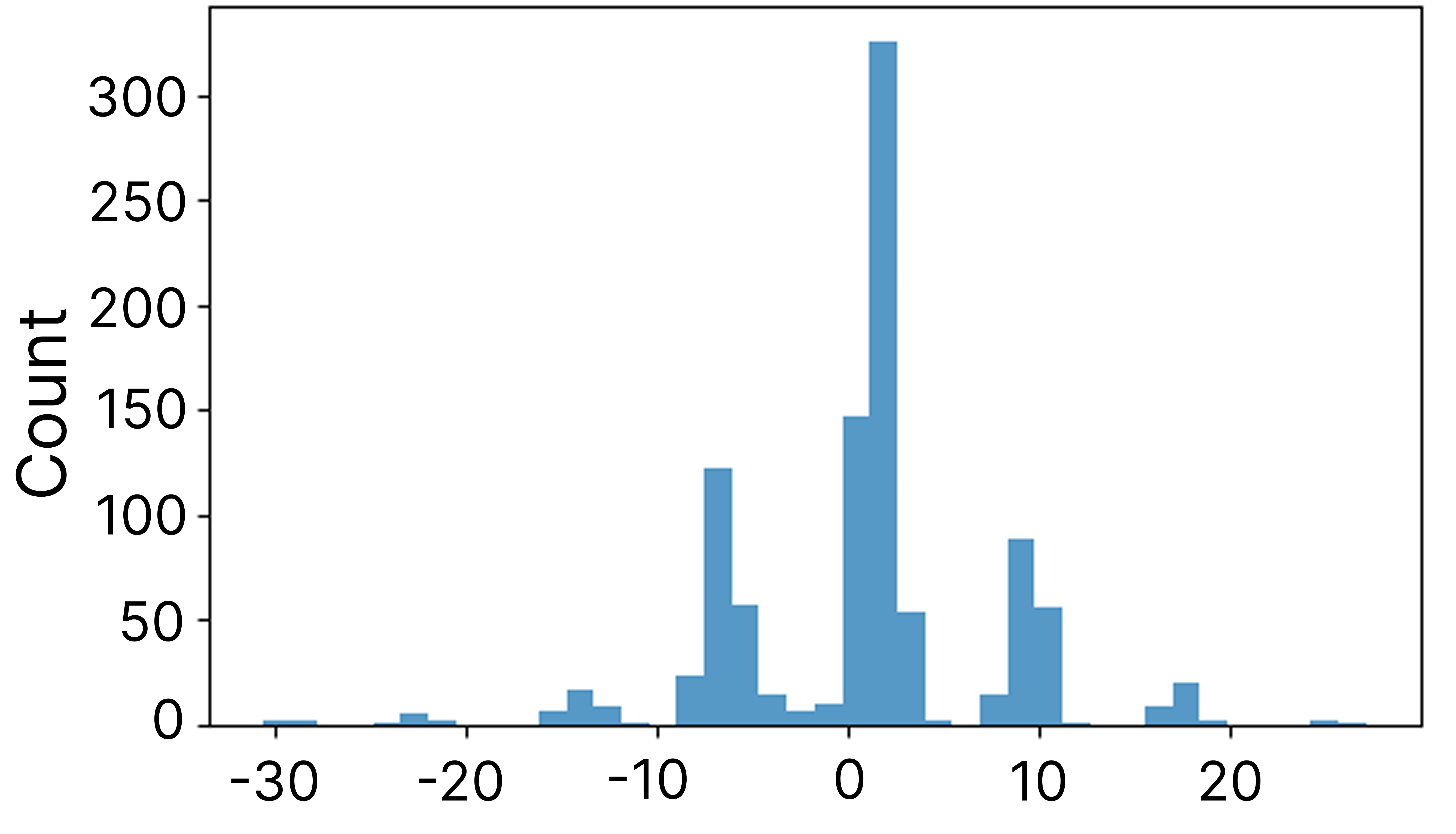}}~\subfigure[Pythia2.8b]{
\label{2.8b_ac_8}
\includegraphics[width=0.4\textwidth]{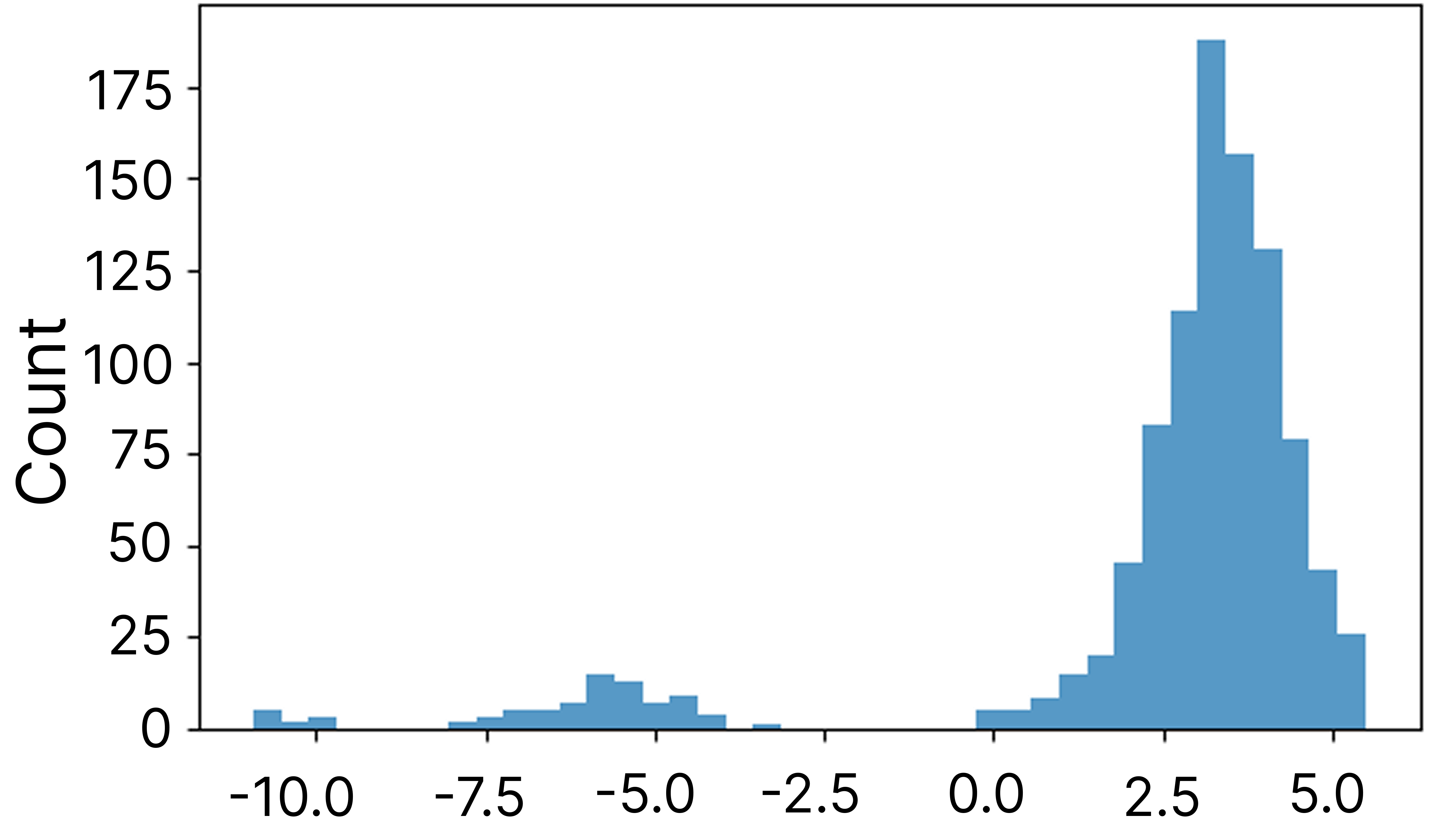}}
\caption{Distribution of log-probability changes after patching the top 20 CoT features into NoCoT runs under dictionary ratio 8. Left: Pythia-70M; Right: Pythia-2.8B. Compared to ratio 4, the distributions are similar: 2.8B continues to show consistent improvements, while 70M remains less robust, exhibiting high variance and frequent negative effects.}
\label{3}
\end{figure*}

\begin{figure*}[h]
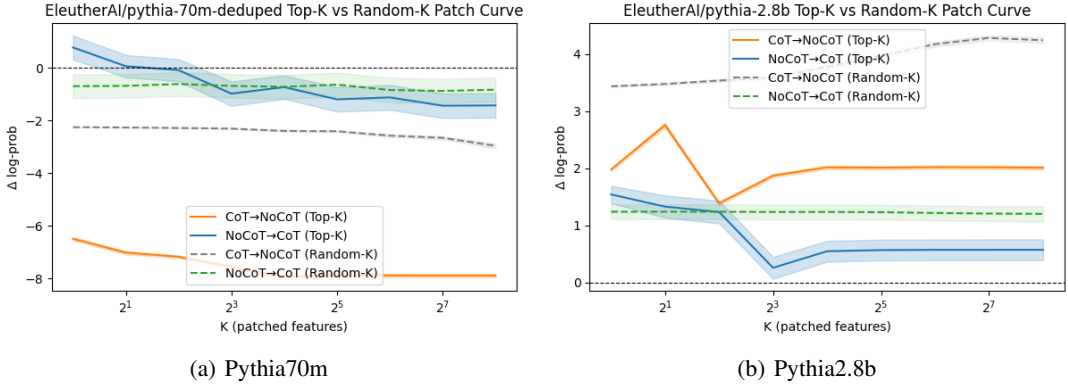

\centering 
\subfigure[Pythia70m]{
\label{70m_pc_4}
\includegraphics[width=0.4\textwidth]{final/70m_ran_4.png}}~\subfigure[Pythia2.8b]{
\label{2.8b_pc_4}
\includegraphics[width=0.4\textwidth]{final/2.8b_ran_4.png}}
\caption{Top-$K$ and Random-$K$ patching performance under dictionary ratio 4. Left: Pythia-70M; Right: Pythia-2.8B. CoT$\rightarrow$NoCoT patching shows the effect of patching CoT features into NoCoT, while NoCoT$\rightarrow$CoT patching shows the reverse. In 2.8B, patching CoT features yields consistent performance gains, highlighting their causal importance. In contrast, for 70M, patching CoT features leads to a substantial and monotonic performance decline, suggesting that CoT-induced features are ineffective or even harmful in the smaller model ($p < 0.001$).}
\label{5}
\end{figure*}

\begin{figure*}[h]
\centering 
\subfigure[Pythia70m]{
\label{70m_pc_8}
\includegraphics[width=0.4\textwidth]{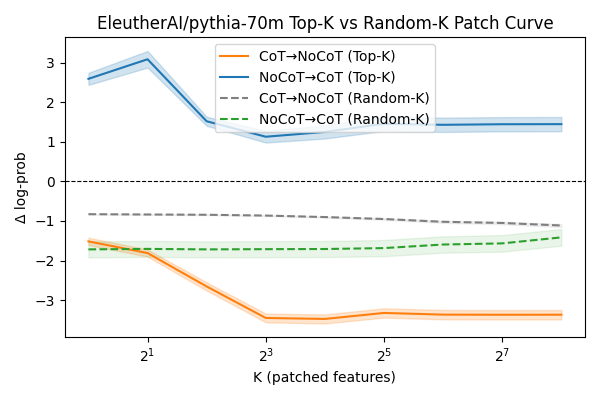}}~\subfigure[Pythia2.8b]{
\label{2.8b_pc_8}
\includegraphics[width=0.4\textwidth]{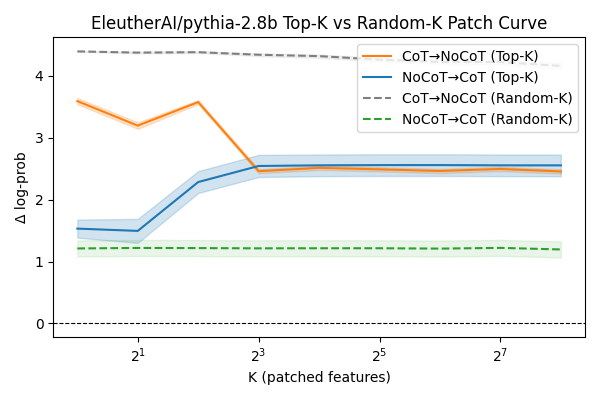}}
\caption{Top-$K$ and Random-$K$ patching performance under dictionary ratio 8. Left: Pythia-70M; Right: Pythia-2.8B. For 2.8B, CoT$\rightarrow$NoCoT patching consistently improves performance, with diminishing returns as $K$ increases. NoCoT$\rightarrow$CoT patching gradually degrades the CoT run, suggesting CoT features are causally significant and sparse. In contrast, for 70M, patching CoT features into NoCoT runs still causes a net performance drop, though less sharply than under ratio 4. Interestingly, NoCoT$\rightarrow$CoT patching shows mild improvement ($p < 0.001$).}
\label{6}
\end{figure*}
We examine the causal role of learned sparse features through a controlled activation patching experiment. Specifically, we keep the model parameters fixed and inject the top-K most salient sparse features from a CoT forward pass into a NoCoT pass, and vice versa, in order to assess their impact on the log-probability assigned to the correct answer.

In the Pythia-2.8B model, we observe a clear directional asymmetry: CoT-to-NoCoT patching tends to improve performance, while NoCoT-to-CoT patching has minimal effect. As shown in Figures \ref{2.8b_ac_4} and \ref{2.8b_ac_8}, this trend holds across both dictionary sparsity ratios of 4 and 8. In each case, the log-probability deltas after patching are predominantly positive, and the distribution is skewed to the right. This indicates that features activated under CoT conditions retain significant causal efficacy even when transferred to NoCoT inputs, effectively "nudging" the model toward more accurate answers.

In contrast, the same patching operation in the smaller Pythia-70M model (Figures \ref{70m_ac_4}) and \ref{70m_pc_8} produces highly unstable results. The effect distribution is nearly symmetric around zero, with positive and negative examples occurring at roughly equal frequencies, and with extreme values (e.g., $\Delta$ log-prob reaching ±30) prominently observed. This suggests that CoT features do not reliably transfer within the smaller model and may even interfere with the original inference trajectory in some cases.

When comparing across dictionary sparsity levels, the pattern remains consistent: both sparsity ratios yield reliable positive transfer effects in Pythia-2.8B, while Pythia-70M consistently shows no stable trend. This supports the view that the observed differences are not artifacts of a particular setup, but instead reflect a broader, capacity-dependent phenomenon—namely, that the causal utility of CoT-derived features is scale-sensitive and more robust in larger models.

To more precisely characterize the relationship between patching performance and the number of patched features $K$, we plot patching curves as shown in Figures \ref{5} and \ref{6}.

We begin with the Pythia-70M model, focusing on the difference between the two patching directions: CoT $\rightarrow$ NoCoT and NoCoT $\rightarrow$ CoT.

Under the dictionary sparsity ratio of 4 (Figure \ref{70m_pc_4}), injecting CoT features into NoCoT trajectories (orange curve) yields no performance gain. In fact, the curve declines steadily after $K > 4$, eventually dropping to around –8 log-prob. This suggests that CoT-activated features may exhibit distributional mismatch or representational conflict in the small model, effectively disrupting the original information processing flow. Conversely, the NoCoT $\rightarrow$ CoT patching (blue curve) also leads to a decline in performance, though the drop is slightly less steep—indicating that the CoT mode may offer some robustness against perturbations.

Under the higher sparsity setting (dictionary ratio = 8), the patching behavior of Pythia-70M continues the trend observed earlier, though the curves appear smoother (see Figure \ref{70m_pc_8}). In the CoT $\rightarrow$ NoCoT direction, the patching curve remains consistently below zero, indicating that features extracted from CoT inputs fail to provide performance gains when injected into NoCoT contexts. In fact, they introduce a degree of disruption to the model's reasoning process. Although the negative impact is numerically less severe compared to the ratio 4 condition (with a minimum drop of about –3, as opposed to –6 to –8), the direction of the effect remains unchanged. This suggests that even under a more relaxed sparsity configuration, the small model is still unable to consistently benefit from the transfer of CoT features.

In contrast, the NoCoT $\rightarrow$ CoT direction reveals a fragile advantage of the CoT setting. At $K=2$, the patching yields a performance boost of approximately +3, suggesting that the first few injected features play a meaningful role in supporting CoT-style reasoning. However, this advantage diminishes rapidly as more NoCoT features are injected, eventually stabilizing around +1 near $K=128$. This trend implies that in small models, CoT-related performance gains may not be driven by a small set of dominant features, but rather distributed across a broader range of components—or that the individual utility of each feature is diluted. As a result, once these features are partially replaced or perturbed, their original advantage becomes difficult to preserve.

Moreover, more random patching experiments also show negative or unstable results. This further supports the idea that CoT-activated features in small models do not transfer well and may cause problems when added to NoCoT trajectories.

Taken together, these observations indicate that Pythia-70M does not successfully encode CoT features with consistent or robust causal influence. Its activation space is more dispersed, and post-patching performance shows high variability, weak directional signal, and susceptibility to disruption.

In contrast, the 2.8B model exhibits a markedly different behavior.

Under dictionary ratio 4 (Figure \ref{2.8b_pc_4}), the orange curve (CoT $\rightarrow$ NoCoT) jumps immediately at $K=2$, reaching a gain of over +2.5 log-prob, then slowly declines to around +1.8—indicating that the top few CoT features carry strong causal weight. In the reverse direction, the blue curve (NoCoT $\rightarrow$ CoT) remains largely flat, showing that replacing features from the CoT pathway has little to no benefit and may even introduce slight interference. 

At a higher sparsity level (ratio 8, Figure \ref{2.8b_pc_8}), this pattern becomes even more pronounced. The orange curve surpasses +3.2 at $K=2$. As $K$ increases, performance slightly declines and then stabilizes at approximately +2.4, revealing a classic "saturation" effect. Meanwhile, the blue curve gradually rises, indicating that NoCoT $\rightarrow$ CoT patching progressively erodes CoT-mode performance. The 2.8B model shows clear performance improvement when transferring from CoT to NoCoT, and significant effects can be observed even with a small number of injected features.

However, after adding Random-K controlled experiments, we find that the performance gains are not due to a specific set of "Top-K strong features." Instead, in the CoT $\rightarrow$ NoCoT direction, randomly selecting K CoT-activated features often leads to better performance than using the Top-K features. This suggests that the useful information activated by CoT prompts is not concentrated in a few highly activated features, but is more widely spread across many moderately activated ones. The Top-K strategy, which only focuses on activation strength, may overfit to local peaks and miss other supportive features that actually play a causal role. In contrast, random sampling is more likely to include these overlooked features, leading to more stable and comprehensive positive effects. We will further explain this phenomenon through an analysis of feature sparsity structure in Section \ref{sec: sparsity}.

Overall, the activation patching experiments confirm that CoT-triggered features exhibit clear causal efficacy in large models: injecting even a small number of CoT-activated sparse features significantly improves model output quality. Interestingly, we find that randomly selected features often outperform top-ranked ones, suggesting that the causal signal is not concentrated in a few dominant directions but rather distributed across a broader feature space. In contrast, the CoT-activated features in small models are more scattered and fragile, lacking stable transferability and in some cases even introducing interference. All patching effects achieved statistical significance ($p < 0.001$), confirming these patterns reflect systematic differences rather than random variation.

Moreover, we observe that the sparsity ratio affects how information is distributed across features. Under higher sparsity (ratio 4), performance gains tend to occur in more abrupt "jumps" but are also more susceptible to outliers. In contrast, with lower sparsity (ratio 8), performance changes are smoother, suggesting more stable and cumulatively effective information transmission.

Together, these results support our central hypothesis: CoT prompting induces a distributed and causally meaningful internal structure, particularly in LLMs where such features are more pronounced and reliably transferable.

\subsection*{Activation Sparsity under CoT and NoCoT}
\label{sec: sparsity}
Following the causal intervention experiments, we now turn to the structural properties of internal activations. In particular, we focus on sparsity—how CoT and NoCoT prompts affect the distribution and density of activated neurons and SAE features. Sparsity is widely associated with interpretability and generalization, and may offer additional insights into the mechanistic impact of CoT.

\begin{figure*}[h]
\centering 
\subfigure[Pythia70m]{
\label{70m_as}
\includegraphics[width=0.4\textwidth]{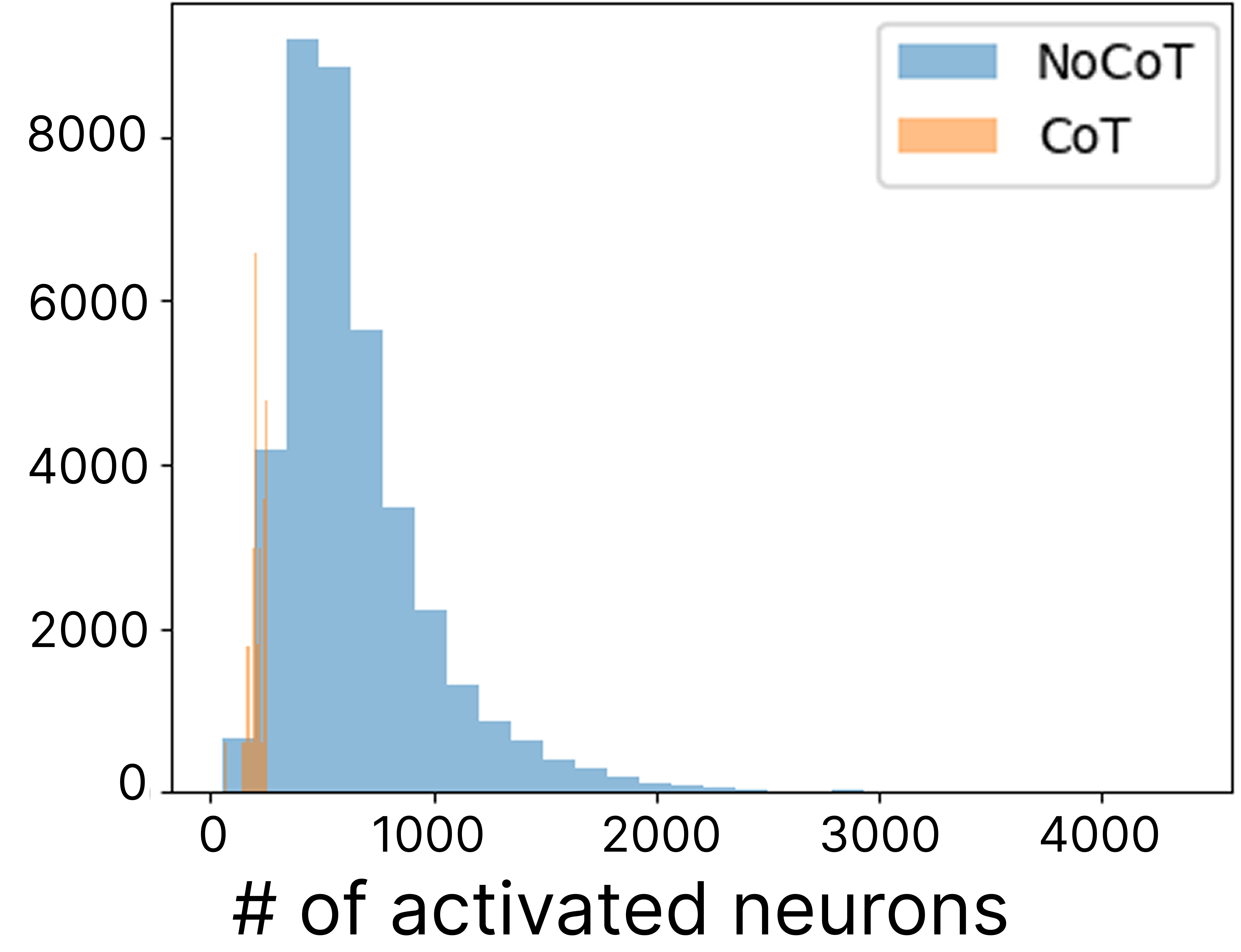}}~\subfigure[Pythia2.8b]{
\label{2.8b_as}
\includegraphics[width=0.4\textwidth]{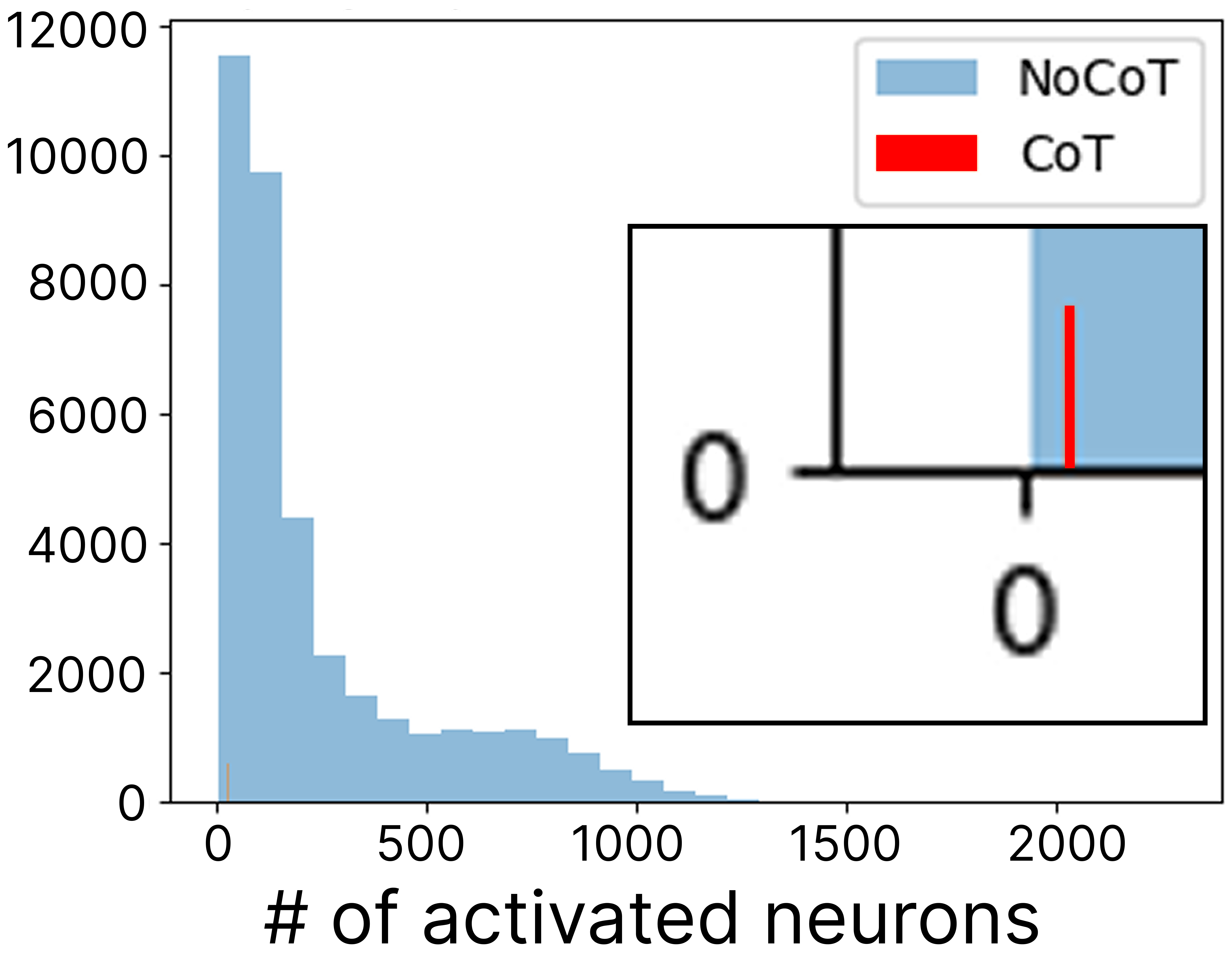}}
\caption{Sparsity comparison of residual activations under CoT and NoCoT prompts. In both models, CoT leads to significantly sparser residual activations, with most neurons remaining near zero and only a small subset strongly activated. This sparsity effect is markedly more pronounced in the 2.8B model, indicating enhanced activation selectivity and structured feature usage at larger scale.}
\label{9}
\end{figure*}

As shown in Figures \ref{9}, we compare the global distribution of residual activations under CoT and NoCoT prompting conditions for the 70M and 2.8B models. The results reveal that CoT prompts lead to significantly sparser residual activations compared to NoCoT prompts. Specifically, in the NoCoT condition, activation values are distributed more broadly, indicating that more neurons exhibit moderate to high activation. In contrast, under CoT prompting, most neuron activations are concentrated in a very low range, with only a few neurons showing strong activation. This sparsity trend appears in both the smaller 70M model and the larger 2.8B model, but is more pronounced in the latter. Notably, in the 2.8B model, the activation distribution under NoCoT has a heavier tail—more neurons exhibit high activation—whereas under CoT, activations are almost entirely low, with only a small subset strongly activated, highlighting a sharper sparsity effect.

\begin{figure*}[h]
\centering 
\subfigure[Pythia70m]{
\label{70m_box_no}
\includegraphics[width=0.4\textwidth]{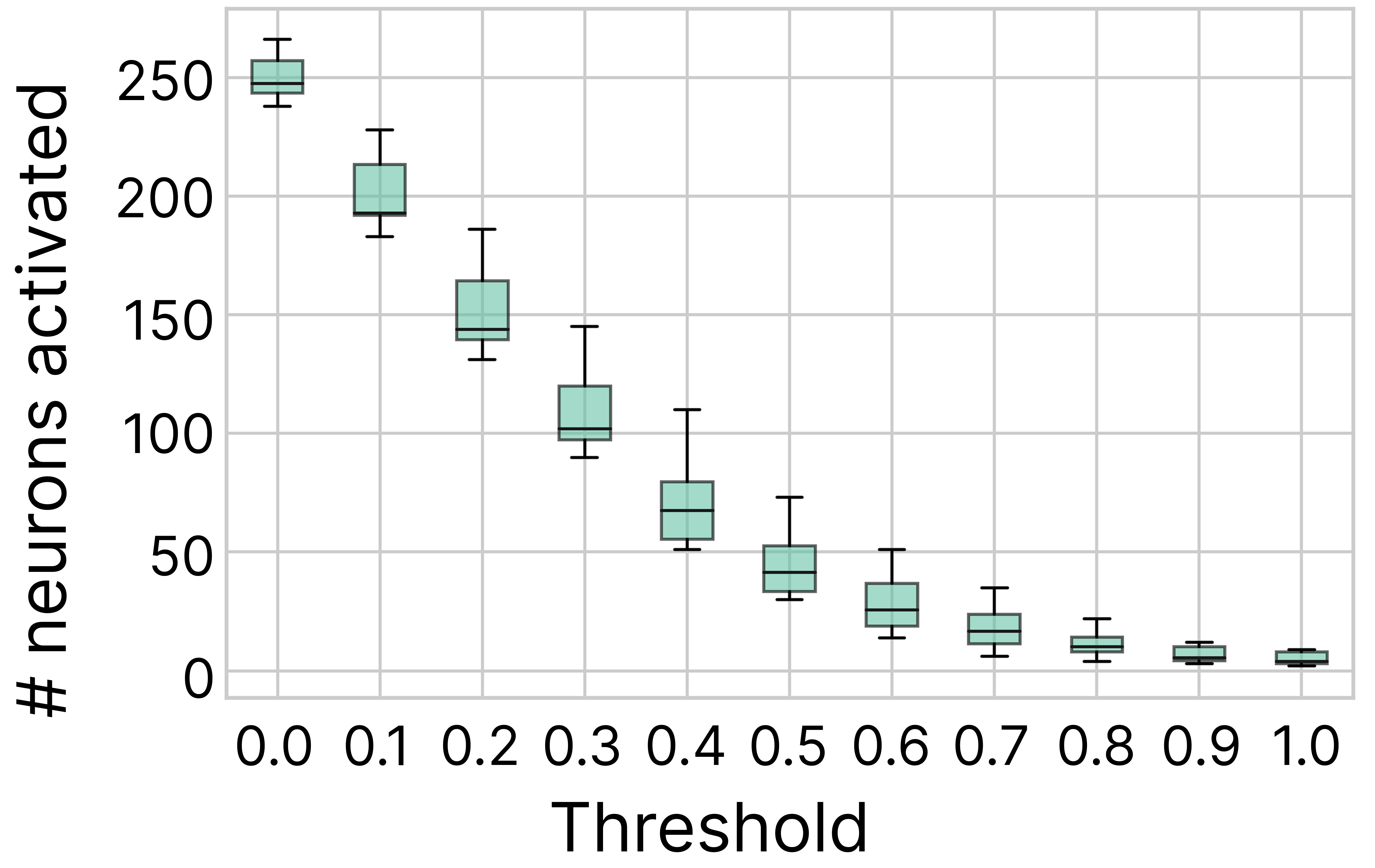}}~\subfigure[Pythia2.8b]{
\label{2.8b_box_no}
\includegraphics[width=0.4\textwidth]{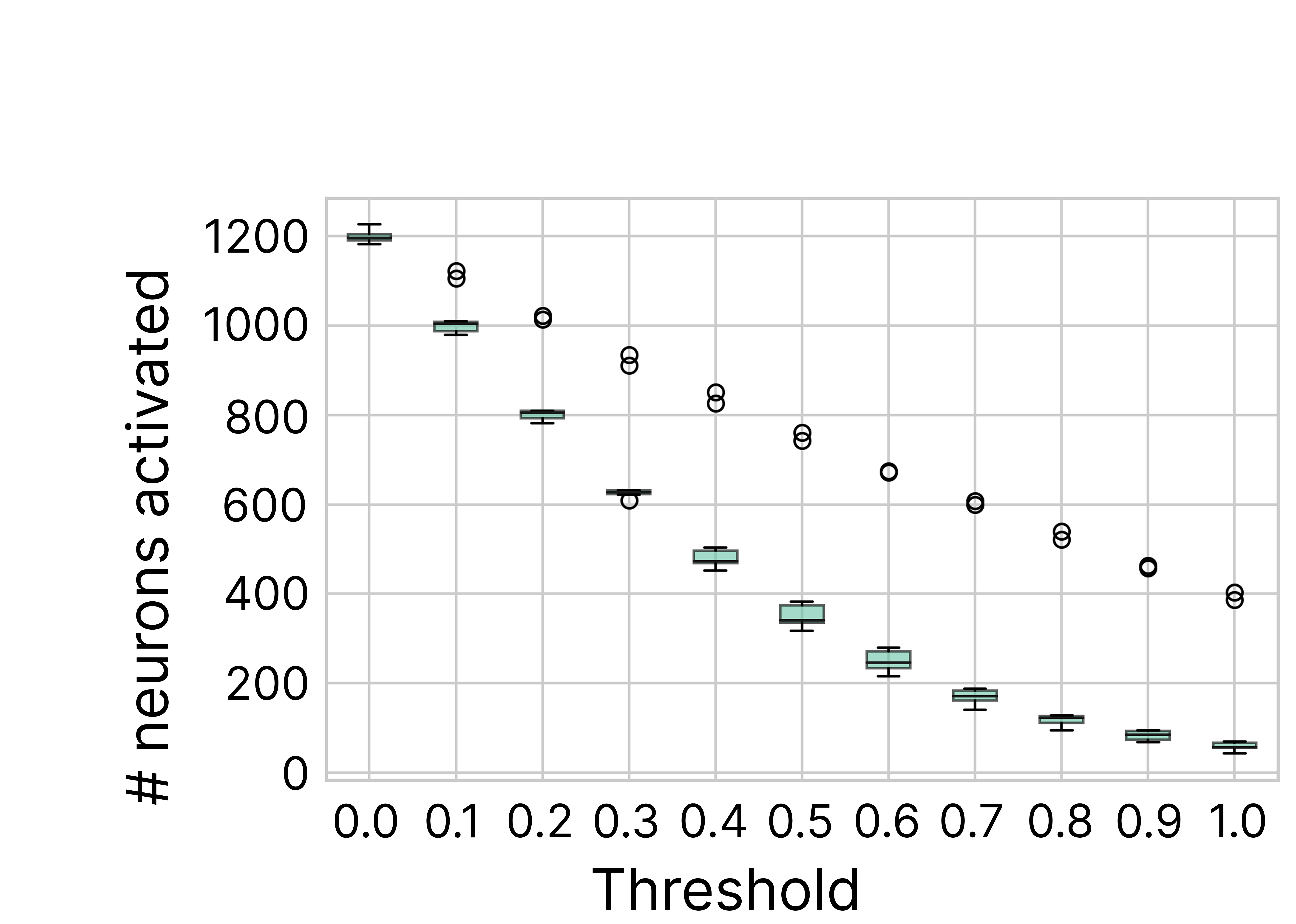}}
\caption{Activated neuron counts per SAE feature under NoCoT prompting, across thresholds from 0.0 to 1.0. The large model (2.8B) activates significantly more neurons per feature at each threshold, indicating denser feature composition compared to the small model.}
\label{7}
\end{figure*}

\begin{figure*}[h]
\centering 
\subfigure[Pythia70m]{
\label{70m_box_cot}
\includegraphics[width=0.4\textwidth]{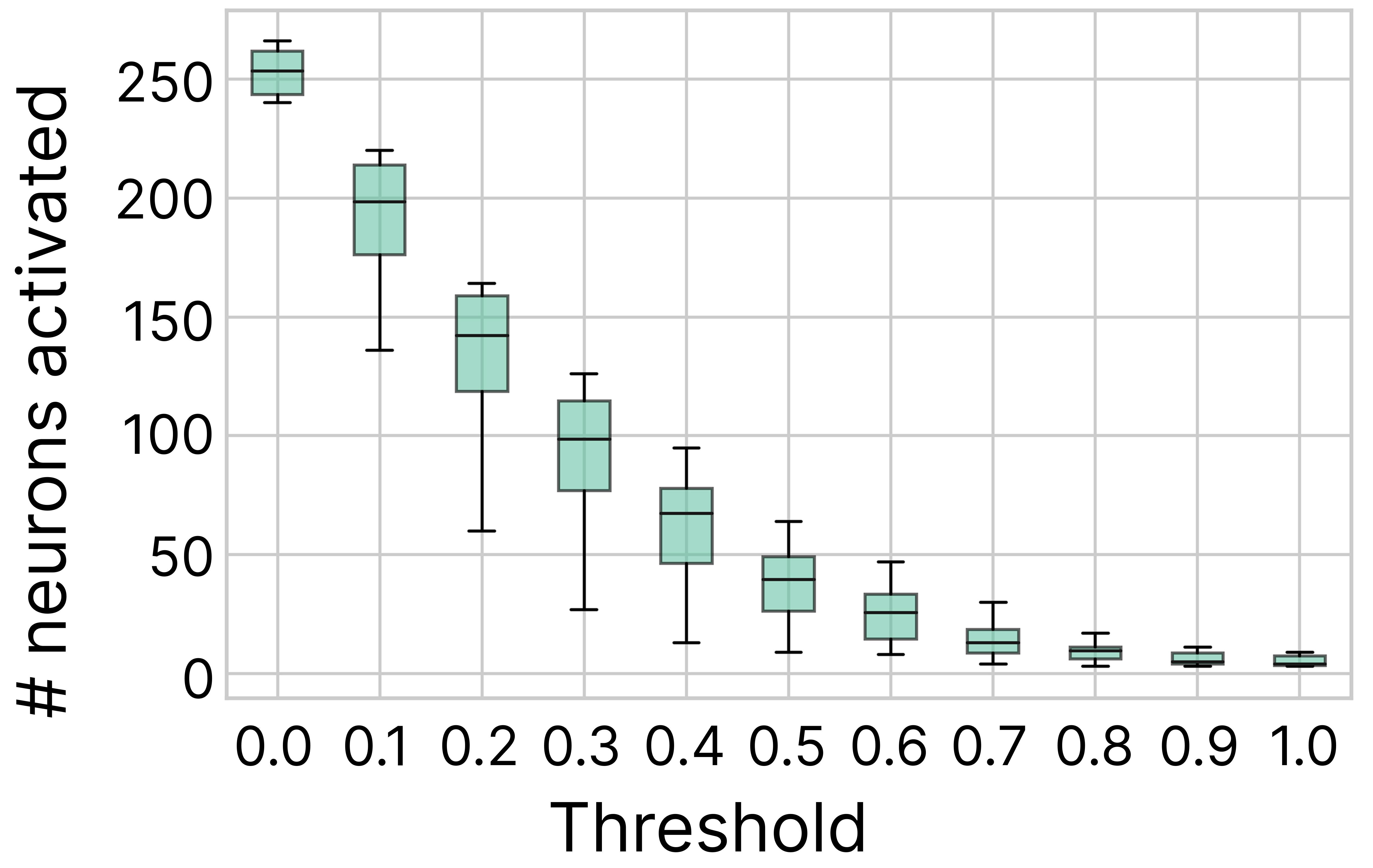}}~\subfigure[Pythia2.8b]{
\label{2.8b_box_cot}
\includegraphics[width=0.4\textwidth]{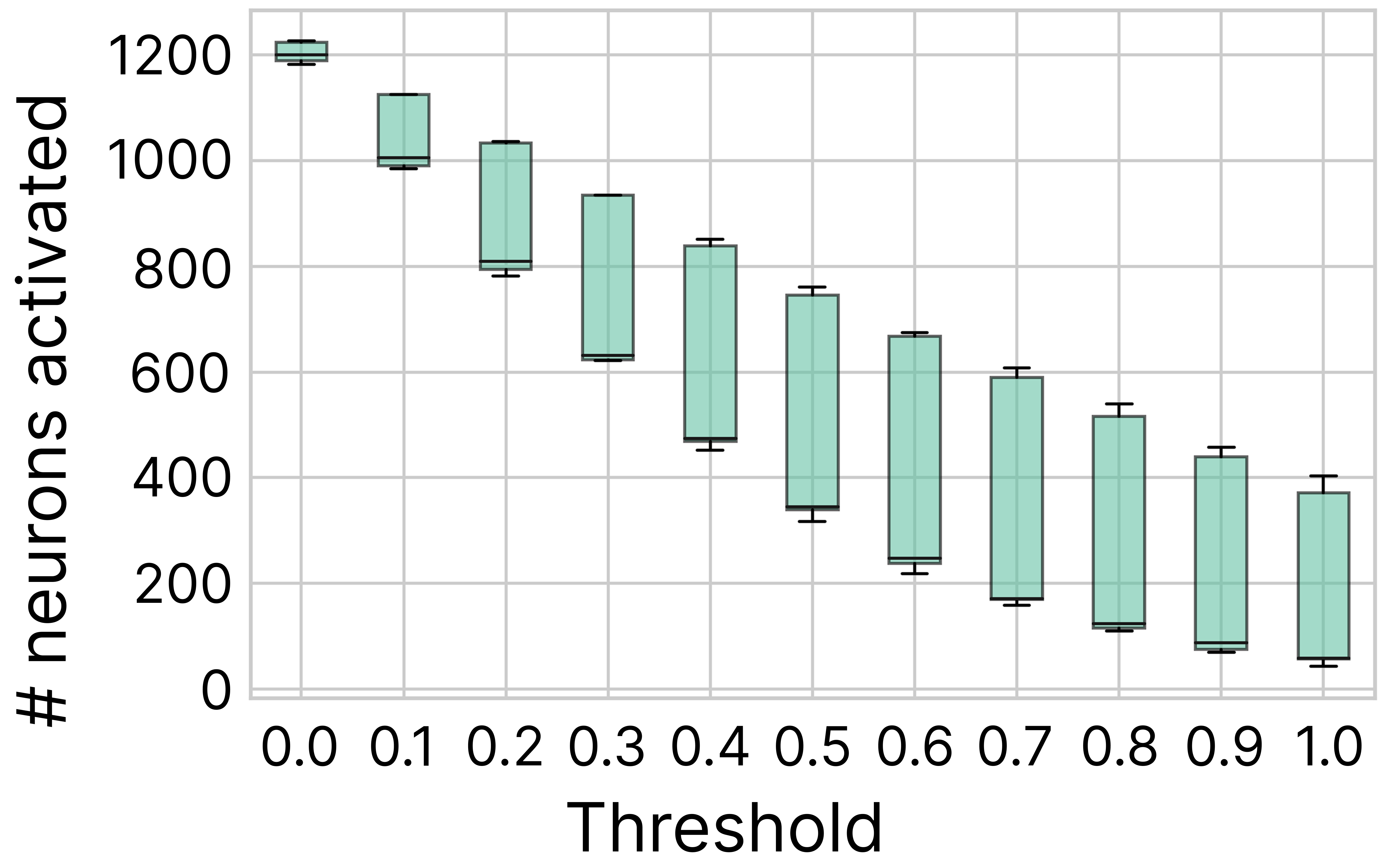}}
\caption{Activated neuron counts per SAE feature under CoT prompting. Compared to NoCoT, CoT prompts yield substantially sparser activations in both models, with 2.8B showing stronger sparsity and higher inter-feature variance.}
\label{8}
\end{figure*}

To further analyze this difference, we apply SAE to extract feature representations from the residual activations, as described in the Methods section, and count the number of significantly activated neurons per SAE feature. Figures \ref{70m_box_no} and \ref{70m_box_cot} show the neuron activation distributions per SAE feature under NoCoT and CoT conditions for the 70M model. A comparison of the two reveals that under CoT, each SAE feature tends to activate only a small number of neurons, whereas under NoCoT, the same features often activate a broader set of neurons. In other words, NoCoT features are associated with more widespread neuron activations, while CoT features are more concentrated and rely on a smaller subset of neurons. This suggests that CoT leads to sparser internal representations in the 70M model, with each feature being encoded by a more compact neuronal subspace.

A similar pattern is observed in the larger 2.8B model, but to a greater extent. Figures \ref{2.8b_box_no} and \ref{2.8b_box_cot} show the SAE feature activation patterns under NoCoT and CoT conditions, respectively. Under NoCoT, each feature still activates a relatively large number of neurons, while under CoT, only a very small subset is strongly activated per feature. Compared to the 70M model, the 2.8B model shows more extreme sparsity: many features are supported by only a handful of neurons, emphasizing that larger models exhibit a more pronounced sparsity trend under CoT and may encode CoT-related features more efficiently.

This phenomenon may seem paradoxical: the CoT activations in 2.8B are globally the sparsest (Figure \ref{2.8b_as}), yet the variance in the number of activated neurons per feature is higher (Figure \ref{2.8b_box_cot}). We interpret this as evidence of a more refined form of structured sparsity in larger models. Rather than uniformly suppressing all features, the large model under CoT appears to allocate representational resources more strategically: some features are highly focused, requiring only a few neurons, while others are more complex and involve broader neuronal collaboration. This increasing divergence in feature-level activation may underlie the superior performance of 2.8B on multi-step reasoning tasks.

Together, these experiments show that CoT prompting not only improves reasoning performance but also reshapes the internal activation patterns of the model. In both 70M and 2.8B, CoT results in fewer neurons being activated overall, indicating greater global sparsity—especially in the 2.8B model. However, this change is not limited to fewer activations: at the SAE feature level, we observe significantly greater variation in how many neurons are engaged by each feature. This suggests that CoT encourages semantic resource allocation, where some features are represented by highly selective neurons and others mobilize a larger population for more complex reasoning. The trend is especially prominent in the 2.8B model, indicating that larger models are not only more sensitive to sparsification, but also more capable of implementing structured sparsity. We argue that this may serve as an indirect mitigation of the superposition problem: by compressing activations and increasing feature separation, CoT prompts induce a form of latent disentanglement. Although this "unsupervised disentanglement" is not explicitly optimized during training, it emerges as a byproduct of semantic prompting and plays a critical role in making internal representations more interpretable and causally effective.

Interestingly, this structured sparsity in CoT-induced representations also helps explain the surprising result from our patching experiments: in the 2.8B model, randomly sampled CoT features consistently outperform top-ranked ones when patched into NoCoT trajectories. At first glance, this seems counterintuitive—why would unranked features yield better performance than those with highest activation?

As shown earlier, CoT prompting not only makes the overall activation in both models more sparse, but also leads to stronger sparsity and higher feature-level variability in the larger model. Specifically, in the Pythia-2.8B model, under CoT conditions, most neurons have their activation values suppressed close to zero, with only a small number being strongly activated. At the same time, the number of neurons involved in different features varies much more. This means that CoT prompts in the large model lead to a form of "structured sparsity": the model does not suppress all features equally, but allocates its limited representational resources more strategically. Some features are highly concentrated and can be represented with only a few neurons, while others, which are more complex, recruit a wider set of neurons to represent them. 

In other words, CoT-related information in the 2.8B model is not carried by just a few strongly activated features, but is spread across combinations of many features. Therefore, simply selecting the Top-K features based on the highest activation values may only cover local peaks in the CoT-related semantics, while missing many moderately activated but still important supporting features. These overlooked features also play a key role in final reasoning, but are not included in the Top-K set. In contrast, when K features are selected randomly, without relying on a fixed ranking by activation strength, there is a higher chance of including these useful but less prominent features. This helps provide a more complete injection of causal information overall. This explains why, in the 2.8B model, the Random-K strategy achieves better CoT $\rightarrow$ NoCoT transfer performance than the Top-K strategy: random sampling covers a richer subset of features and avoids focusing too narrowly on local activation peaks, allowing it to capture more useful signals.

In contrast, this phenomenon does not appear in the 70M model. One possible reason is that the difference in feature distributions between CoT and NoCoT conditions is much smaller compared to the larger model. In the small model, CoT prompting does increase activation sparsity to some extent, but the overall feature activation patterns remain similar to those under NoCoT. For example, in Pythia-70M, each sparse feature under CoT typically activates only a small number of neurons, while the same feature under NoCoT might activate a wider set of neurons. However, this feature-level sparsification is much weaker than what is observed in the 2.8B model. More importantly, the limited capacity of the 70M model makes it difficult to develop new internal structures or feature organization patterns in response to CoT prompts. As discussed earlier, CoT does not significantly improve the interpretability or consistency of features in the 70M model. As a result, there are not many additional useful features emerging under CoT that the model can take advantage of. Both Top-K and Random-K strategies end up inserting features that are similarly noisy or irrelevant to the model, which naturally leads to no clear difference in performance or consistent gains. This also aligns with our earlier conclusion: smaller models, due to their limited representational power, are less capable of capturing and using the structured reasoning signals introduced by CoT prompting, and show very limited improvements in the causal relevance of their internal activations.

These results connect the earlier patching experiments with the structural analysis in this section. They show that CoT prompts create sparse, disentangled, and compositional representations in larger models, making it easier to replace features and maintain reasoning quality. In contrast, small models lack this structure, which limits their ability to benefit from CoT-style prompting. This supports the main idea that model size is critical for making CoT-induced features causally effective and well-organized.

\section*{Code and Reproducibility}
To support reproducibility, we provide the code, configuration files, and experiment instructions in an anonymous GitHub repository:

\url{https://github.com/sekirodie1000/cot_faithfulness}

\end{document}